\documentclass[10pt,journal]{IEEEtran}
\ifCLASSINFOpdf
\else
\fi
\usepackage{hyperref}
\usepackage{amsmath}
\usepackage{amsfonts}
\usepackage{amssymb}
\usepackage{graphicx}
\usepackage{subcaption,wrapfig}
\usepackage{epstopdf}
\usepackage{multirow}
\usepackage{cite}
\usepackage{enumitem}
\usepackage{makecell}
\usepackage{color}
\usepackage[ruled,vlined,linesnumbered,noresetcount]{algorithm2e}
\usepackage{algorithmic}

\begin{document}
%
\title{Rate-Adaptive Neural Networks for \\Spatial Multiplexers}
%
%
%

\author{Suhas~Lohit, Rajhans~Singh,
        Kuldeep~Kulkarni~and~Pavan~Turaga
\IEEEcompsocitemizethanks{\IEEEcompsocthanksitem S. Lohit, R. Singh and P. Turaga are with the Department of Electrical, Computer and Electrical Engineering and the Department of Arts, Media and Engineering, Arizona State University, Tempe, AZ\protect\\
E-mail: slohit@asu.edu, rsingh70@asu.edu, pturaga@asu.edu
\IEEEcompsocthanksitem K. Kulkarni is with the Department of  Electrical and Computer Engineering, Carnegie Mellon University, Pittsburgh, PA \protect\\
E-mail: kuldeepk@andrew.cmu.edu \protect\\
}
}

\maketitle

\begin{abstract}
In resource-constrained environments, one can employ spatial multiplexing cameras to acquire a small number of measurements of a scene, and perform effective reconstruction or high-level inference using purely data-driven neural networks. However, once trained, the measurement matrix and the network are valid only for a single measurement rate (MR) chosen at training time. To overcome this drawback, we answer the following question: How can we jointly design the measurement operator and the reconstruction/inference network so that the system can operate over a \textit{range} of MRs? To this end, we present a novel training algorithm, for learning \textbf{\textit{rate-adaptive}} networks. Using standard datasets, we demonstrate that, when tested over a range of MRs, a rate-adaptive network can provide high quality reconstruction over a the entire range, resulting in up to about 15 dB improvement over previous methods, where the network is valid for only one MR. We demonstrate the effectiveness of our approach for sample-efficient object tracking where video frames are acquired at dynamically varying MRs. We also extend this algorithm to learn the measurement operator in conjunction with image recognition networks. Experiments on MNIST and CIFAR-10 confirm the applicability of our algorithm to different tasks.
\end{abstract}


%
\IEEEpeerreviewmaketitle

\section{Introduction}\label{sec:introduction}
%
%
%
%
\IEEEPARstart{W}{ith} the help of neural networks and large datasets, researchers have shown significant improvements in performance in high-level inference problems like image recognition as well as low-level inverse problems in vision. Over the last few years, neural networks have been shown to provide state-of-the-art image reconstructions from random Gaussian measurements in real-time \cite{mousavi2015deep,Kulkarni_2016_CVPR}. These results have been further improved by jointly learning the measurement matrix and the reconstruction network \cite{mousavi2015deep,lohit2017convolutional}. Other related work includes those that employ deep learning for reconstruction of CS videos \cite{iliadis2018deep,kai2017csvideonet} and learning sensor multiplexing \cite{chakrabarti2016learning}. These methods are very effective at recovering image estimates from a small set of linear projections of the scene obtained using a spatial multiplexing camera (discussed later in section \ref{sec:related_work}). It has also been shown that if inference is the intended application, it can be performed directly in the compressed domain with reasonable success \cite{calderbank2009compressed,Davenport}. Such direct inference from random Gaussian measurements, i.e., inference without reconstructing the scene, can be perfomed more effectively with neural networks \cite{lohit2016direct}, and can be further improved by jointly learning the measurement operator and the inference network \cite{CL_DNN_2016}. These works demonstrate that spatial-multiplexers can be used to acquire very few measurements (sometimes just 1\% of the measurements) without sacrificing inference performance. Over the last decade, numerous algorithms for spatial multiplexers have been developed and are useful in application domains where constraints on bandwidth, memory and energy are the main bottlenecks, such as in mobile devices and autonomous drones.

\begin{figure*}[t]
  \centering
  \includegraphics[trim = {3cm, 10cm, 3.5cm, 10.8cm}, clip, width=\textwidth]{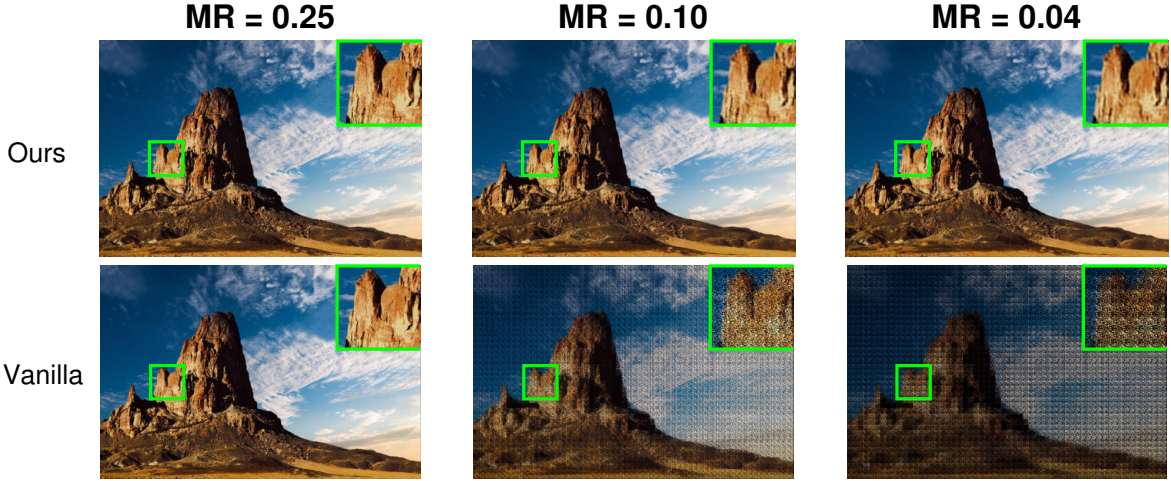}
\vspace{-0.7in}
\caption{Sample visualizations comparing our algorithm Rate-Adaptive CS which is stable over a range measurement rates $[0.04-0.25]$, with vanilla training algorithm that is trained for a single MR=0.25 (272 measurements) and tested over the chosen MR range. Clearly, our algorithm produces superior quality images over the entire range of MRs.}
\label{fig:examples}
\end{figure*}

\begin{figure*}[t]
\centering
\includegraphics[trim = {1cm, 1cm, 1cm, 1cm} ,clip,width=\textwidth]{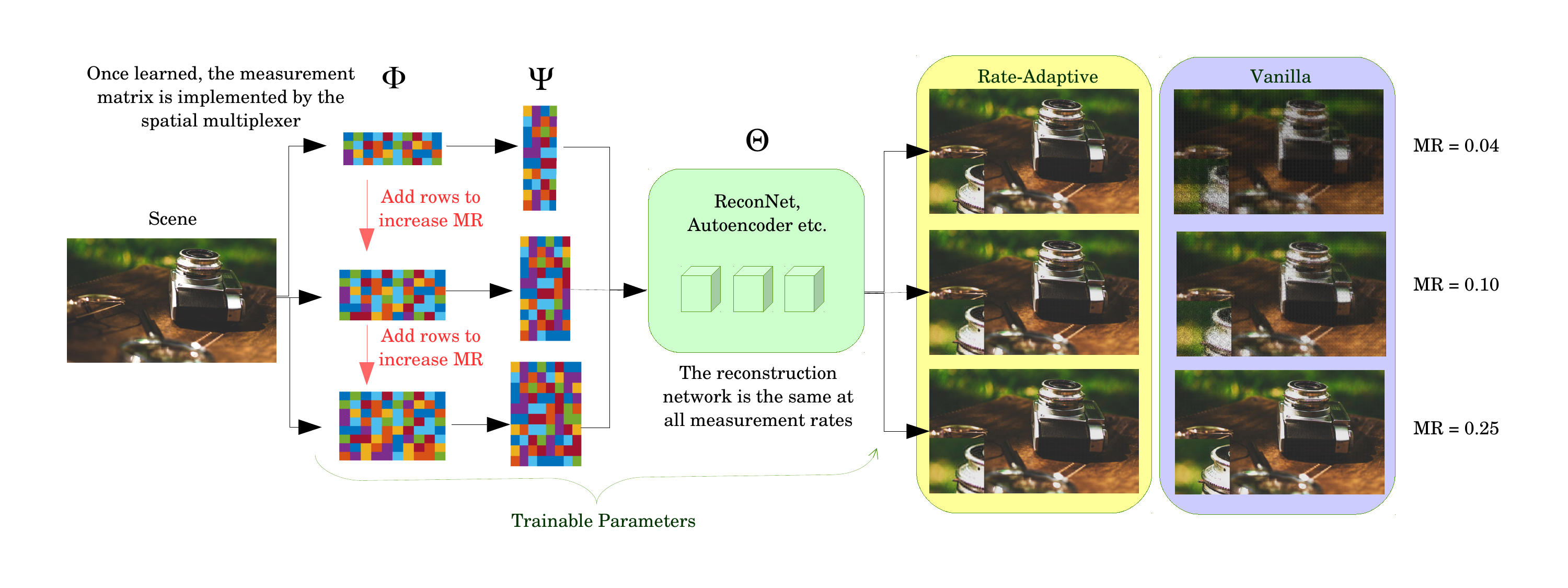}
\caption{The figure shows Rate-Adaptive CS for image reconstruction. The measurement matrix, $\Phi \in \mathbb{R}^{m \times n}$ forms the first FC layer in the network. After training, $\Phi$ is implemented in the SPC, and the output of this layer are the compressive measurements of the scene. The second layer $\Psi \in \mathbb{R}^{n \times m}$ maps the compressive measurements back into the 2D space. The rest of the reconstruction network (ReconNet, autoencoder, DR\textsuperscript{2}-Net etc.) is represented by $\Theta$. For image recognition, we use an inference network instead.}
\label{fig:learning_mm}
\end{figure*}

The ratio of the number of multiplexed measurements acquired to the number of pixels in the reconstructed image is called the measurement rate (MR).\textbf{ A common feature of the deep learning-based works mentioned above is that, for a given trained network, the MR is defined prior to training, and thus, cannot be used at different MRs at test time. In this paper, we extend the ideas of deep learning based data-driven CS reconstruction and inference to applications where it is necessary and useful to allow for the MR to vary at test time. We call such a system -- Rate-Adaptive CS}, and it refers to the combination of the multiplexer (implemented by the camera) and the reconstruction/inference network that follows it. Note that both these components are learned jointly in an end-to-end deep learning framework. This is illustrated in Figure \ref{fig:learning_mm}. As practical applications of Rate-Adaptive CS, one can envision a number of power and storage-constrained mobile systems, where one would like MR to be a function of available battery, storage, or time-varying bandwidth constraints or even content-based dynamically varying MR. This also means that only a single n/w needs to be stored in the system and no access to training data is necessary. Hence, our approach is memory efficient, compared to earlier purely data-driven deep-learning-for-CS methods.

For the task of image reconstruction, compared to earlier approaches like Lohit et al. \cite{lohit2017convolutional}, we get huge improvements in PSNR of up to 15 dB, for the case wherein a network trained at a MR = 0.25 is used to reconstruct measurements acquired at MR = 0.04. Some examples are shown in Figure \ref{fig:examples}. All networks are trained using Tensorflow, employing auto-differentiation.

\subsection{Contributions}
\begin{enumerate}
\item We design a novel three-stage training algorithm that allows learning the measurement operator and the reconstruction/inference network jointly such that the system can operate over a range of measurement rates, without any need for further fine-tuning.
\item Using state-of-the-art network architectures for purely data-driven CS reconstruction, we demonstrate experimentally that a single network trained with our algorithm produces high quality image reconstruction for a large range of MRs. We conduct experiments showing the method is valid across different architectures -- CNNs, as well as autoencoders.
\item In practice, MR is a function of several constraints such as energy, memory and bandwidth, which are usually time-varying. We discuss such applications where Rate-Adaptive CS may become necessary. As a proof of concept, we describe object tracking where MR is varied dynamically depending on image content or a pre-determined adaptation scheme. We use a single network to reconstruct the frames at different MRs and show that the object tracking performance remains competitive. 
\item In addition to image reconstruction, this algorithm is applicable to high-level inference tasks, and completely bypasses reconstruction. That is, the algorithm allows for learning rate-adaptive inference networks for tasks like image recognition rather than reconstruction. Experiments on MNIST (LeNet) and CIFAR-10 (ResNet) show the wide applicability of our algorithm. 
\end{enumerate}

\section{Background and Related work}\label{sec:related_work}
The signal acquisition model is $\mathbf{y} = \Phi\mathbf{x}$, where $\mathbf{y} \in \mathbb{R}^m$, $\mathbf{x} \in \mathbb{R}^n$, with $m << n$. Elements of $\mathbf{y}$ are the compressive/multiplexed measurements and $\mathbf{x}$ is the scene. 

\subsection{CS Image Reconstruction Using Neural Networks:}
For the reconstruction task, the goal is to recover $\mathbf{x}$ from $\mathbf{y}$. Since $m < n$, we have an underdetermined linear system and there is no unique solution, in general. Therefore, traditional iterative algorithms exploited the sparsity/compressibility of $\mathbf{x}$ in transform domains e.g. \cite{needell2009cosamp} or learned a model for $\mathbf{x}$, which is then used as a prior to obtain a unique reconstruction \cite{baraniuk2010model}. Recently, deep learning algorithms were shown to be significantly superior to these conventional approaches. Many deep learning approaches are a combination of the iterative algorithms and data. The deep networks either implement the unrolling of iterative algorithm as layers in the network \cite{gregor2010learning,kamilov2016learning,borgerding2016onsager,tramel2016approximate,metzler2017learned,dave2017compressive}, or the network performs one of the steps in the iterations such as the proximal operation \cite{chang2017one}. Although effective, they do not allow for learning the spatial multiplexing patterns jointly with the reconstruction networks, which is the main focus of this paper. Hence, we will consider only purely data-driven approaches that employ deep learning to learn the inverse mapping from $\mathbf{y}$ to $\mathbf{x}$ based on just training data. A data-driven approach may also allow the algorithm to learn more complex patterns from data that may not be easily expressible in a model-based approach.

Learning a neural network has the added advantage of being non-iterative, amenable to parallelization on a GPU and has been shown to perform reconstruction of a $256 \times 256$ image in real-time which is 3 orders of magnitude faster than the earlier iterative methods. The first work was by Mousavi et al. \cite{mousavi2015deep} which used a Stacked Denoising Autoencoder (SDA) to perform the signal recovery. This was followed by ReconNet which is based on a Convolutional Neural Network (CNN) architecture \cite{Kulkarni_2016_CVPR}, inspired by the Super-Resolution CNN (SRCNN) by Dong et al.\cite{dong2016image}. Yao et al.\cite{yao2017dr} recently proposed a modification of the ReconNet architecture with residual connections called DR\textsuperscript{2}-Net, which improves reconstruction results by about 1-2 dB. These works are based on using a fixed Gaussian measurement matrix. Mousavi et al.\cite{mousavi2015deep} also describe an SDA based architecture where both the measurement operator and the reconstruction network are learned jointly. Lohit et al. \cite{lohit2017convolutional} extend the ReconNet architecture with another fully connected layer which serves as the measurement matrix that is programmed into the SPC. We too focus on the joint learning framework, but with design constraints on the system for rate-adaptivity. In this paper, we consider two choices for the underlying reconstruction architecture -- the extended ReconNet architecture \cite{lohit2017convolutional} and an autoencoder, based on \cite{mousavi2015deep} for our experiments. We believe similar trends will be observed with the other purely data-driven methods as well.	

\subsection{Direct Inference on CS Measurements Using Neural Nets}
For the inference task, the goal is to perform inference such as image recognition directly from the compressive measurements without reconstructing the image first. Calderbank et al.\cite{calderbank2009compressed} have shown theoretically that we can learn linear classifiers like SVM directly in the compressed domain that perform nearly as well as in the original domain. Davenport et al.\cite{Davenport} proposed the ``smashed filter", which is match filtering directly in the compressed domain. Naturally, the problem of direct inference has a deep learning solution first shown by Lohit et al.\cite{lohit2016direct}. The idea is to first project the compressive measurements back into the pixel space using $\hat{\mathbf{x}} = \Phi^T\mathbf{y}$, where $\Phi$ is a predefined Gaussian/Bernoulli measurement matrix, and using $\hat{\mathbf{x}}$ as the input to an image recognition network. This yields excellent results compared to the linear correlation filtering techniques. This was later extended by Adler et al.\cite{CL_DNN_2016} to the case where both the measurement matrix and the image recognition network are learned jointly. This further improves results especially at lower measurement rates. We train networks with the latter architecture in order to make them adaptive to varying MR.

\subsection{Spatial multiplexers and compressive imaging}
Advances in computational imaging have led to cameras tailored to application domains. Indeed, different constraints arise depending on the task at hand which necessitate design and exploration of novel camera architectures. For example, mobile devices have energy constraints, imaging in short-wave infrared (SWIR) domain in high resolution is very expensive with conventional sensors, and magnetic resonance imaging is very slow when sampled at the Nyquist rate. Also, when the intended application of the image acquisition is high-level inference, depending on the inference algorithm, pixel intensities may not be the optimal (for e.g., in terms of energy, sensing and communication costs) representation that the camera should generate for a scene. We can overcome these issues using spatial multiplexing cameras, where, instead of recording pixel intensities, projections of the scene onto a chosen basis subset are computed by the camera. Some of the earlier works in this area includes the work by Neifeld and Shankar \cite{neifeld2003feature} who propose `feature specific imaging', where, the images are captured directly in the required task-specific basis e.g. PCA basis or wavelet basis. 

A related framework for imaging based on compressive sampling theory, called compressive imaging, has been developed as a promising signal acquisition paradigm for sampling scenes at sub-Nyquist rates. The single-pixel camera (SPC) is the most popular example of a compressive imager \cite{wakin2006architecture}. It employs a digital micro-mirror array to encode the spatial multiplexing patterns (usually Bernoulli/Gaussian random vectors). A single photo-diode sensitive to the required wavelengths is used as to record the measurements. In the case of SWIR imaging, this greatly reduces the cost of the sensor. Miniaturized compressive imagers have also been developed and are more energy-efficient than conventional imagers \cite{oike2013cmos}. There are variations and improvements of the SPC. These include the block SPC architecture by Kerviche et al.\cite{kerviche2014information} which captures measurements of the scene by splitting into blocks and thus is more easily scalable to larger image resolutions, the P2C2 architecture that uses spatio-temporal priors for reconstruction that enables high-speed imaging of video frames \cite{reddy2011p2c2} and the LiSens architecture by Wang et al.\cite{wang2015lisens}, which instead of a single photodiode employs a line sensor of many photodiodes to speed up the measurement process. 

The aforementioned camera architectures, are based on the theory of compressive sensing (CS) \cite{candes2006robust,donoho2006compressed}, which states that signals which are sparse/compressible in some basis, can be sampled at sub-Nyquist rates and  reconstructed nearly perfectly using random projections to compute the measurements. However, until recently, the plethora of reconstruction algorithms, developed either based on sparsity of images in transform domains or other models, were all iterative in nature and computationally inefficient -- much slower than real-time even for moderately sized images ($256 \times 256$). These algorithms also do not perform well at very low measurement rates (MR $< 0.1$), where the advantages of spatial multiplexing are most evident.

\section{Learning a Single Network for a Range of MRs: Rate-Adaptive CS}\label{sec:Rate-Adaptive CS}
This section describes the main contribution of the paper. As mentioned earlier, it is of practical value if we can train the combination of the measurement matrix and the reconstruction/inference network such that the system can operate at multiple measurement rates, without requiring any further training or fine-tuning. To this end, we propose a new training algorithm that makes the system performance \textbf{\textit{stable}} with respect to a chosen range of measurement rates (MR). By this we mean the following: For a given range of MRs, we want the performance of the algorithm to be highest at the upper limit of the range and decrease slowly as the MR is decreased, such that the performance of Rate-Adaptive CS at any particular MR in the range is approximately equal to that of a system that is trained for a single MR. We call such a system Rate-Adaptive CS. This is an interesting design constraint that has not been considered in the literature of either compressive sensing or deep learning. 

\begin{figure*}[ht!]
\begin{minipage}[c][4cm][t]{.25\textwidth}
  \centering
  \includegraphics[trim = {12cm, 11cm, 15cm, 10cm}, clip,width=\textwidth]{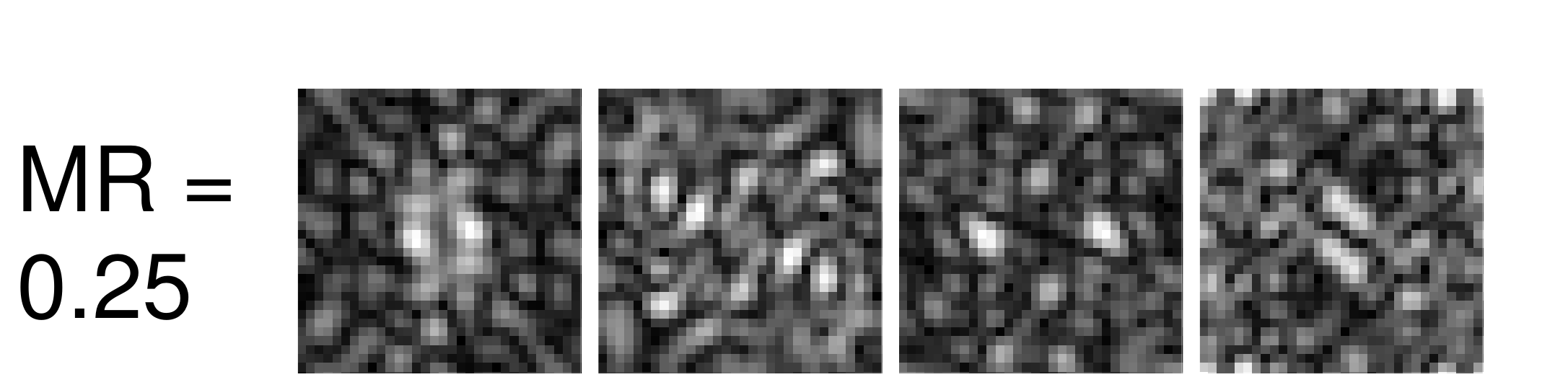}
  \label{fig:test1}
  \centering
  \includegraphics[trim = {12cm, 11cm, 15cm, 10cm}, clip,width=\textwidth]{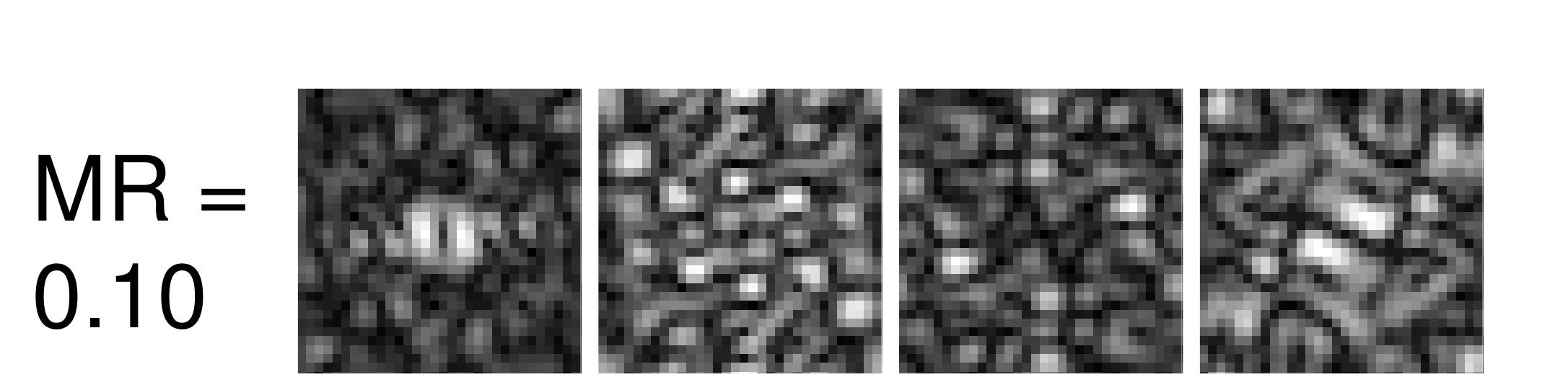}
  \subcaption{conv 1 filters}
  \label{fig:test1}
\end{minipage}%
\begin{minipage}[c][4cm][t]{.25\textwidth}
  \centering
  \includegraphics[trim = {12cm, 11cm, 15cm, 10cm}, clip,width=\textwidth]{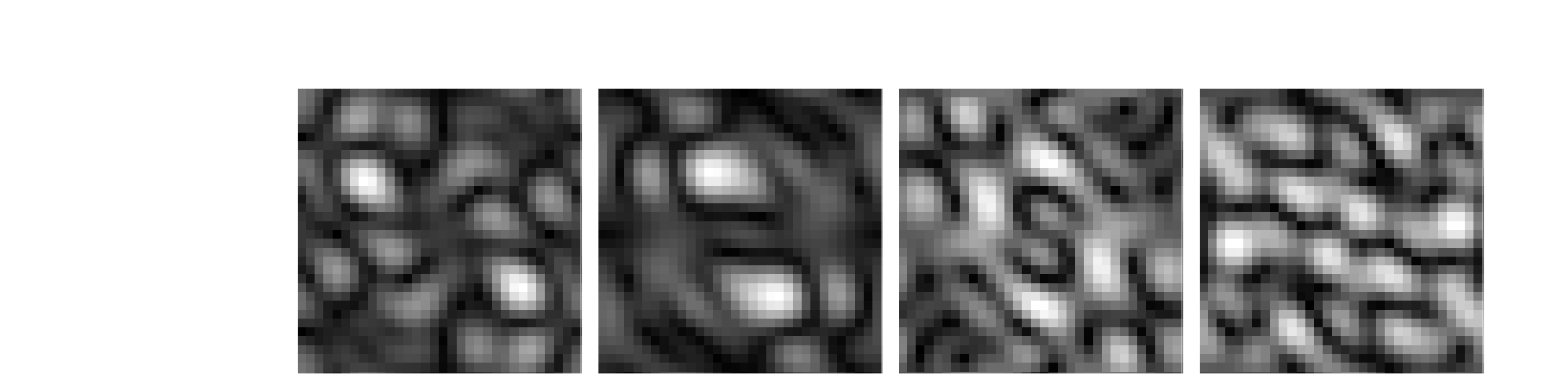}
  \label{fig:test2}
  \centering
  \includegraphics[trim = {12cm, 11cm, 15cm, 10cm}, clip,width=\textwidth]{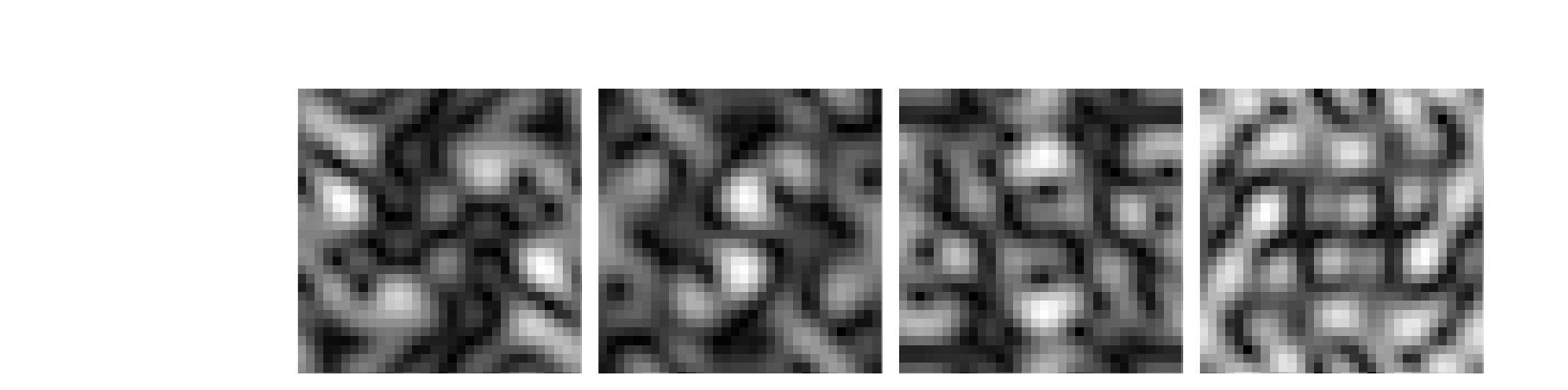}
    \subcaption{conv 3 filters}
\end{minipage}%
\begin{minipage}[c][4cm][t]{.25\textwidth}
  \centering
  \includegraphics[trim = {12cm, 11cm, 15cm, 10cm}, clip,width=\textwidth]{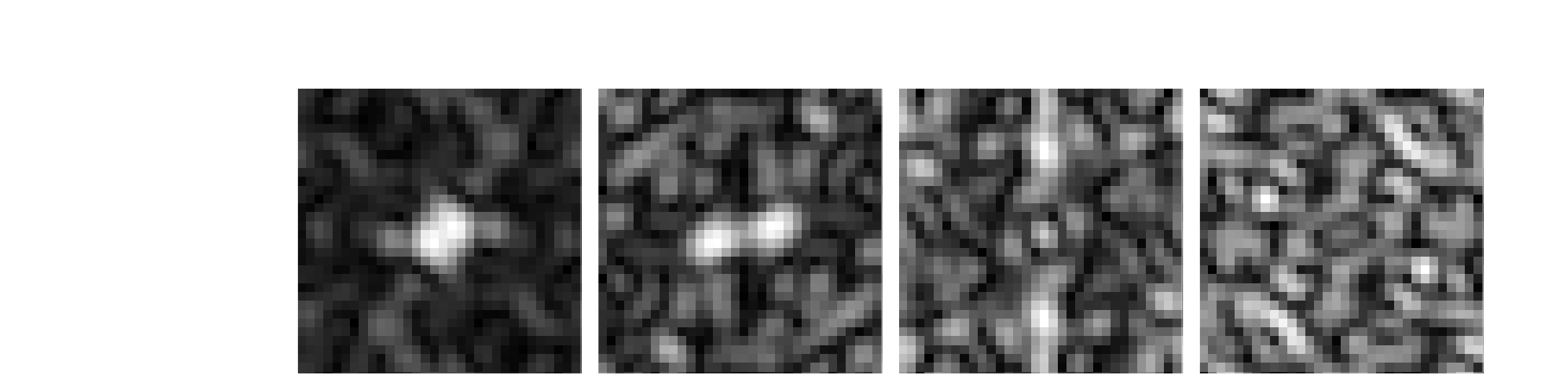}
  \label{fig:test1}
  \centering
  \includegraphics[trim = {12cm, 11cm, 15cm, 10cm}, clip,width=\textwidth]{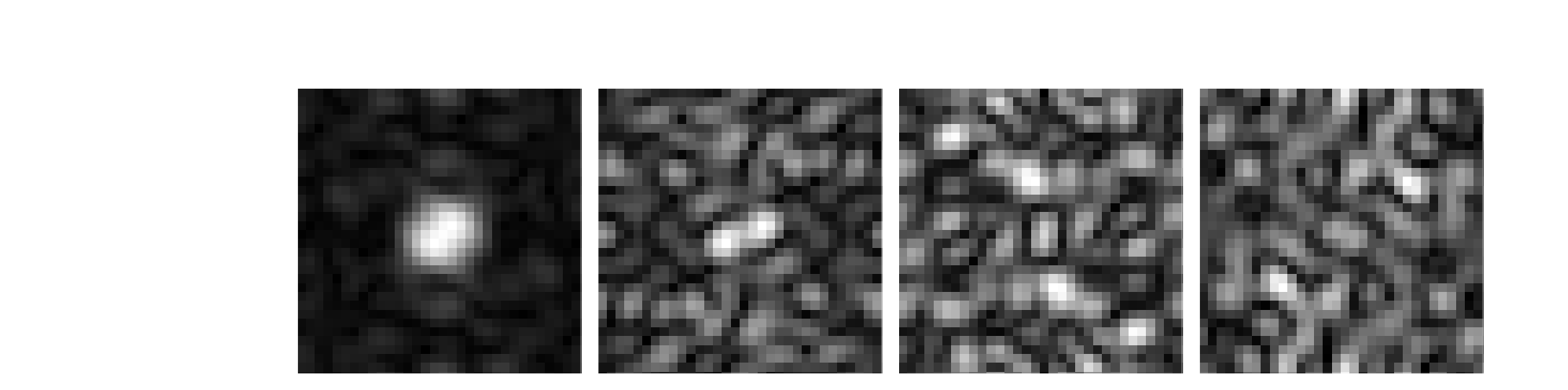}
  \subcaption{conv 4 filters}
\end{minipage}%
\begin{minipage}[c][4cm][t]{.25\textwidth}
  \centering
  \includegraphics[trim = {12cm, 11cm, 15cm, 10cm}, clip,width=\textwidth]{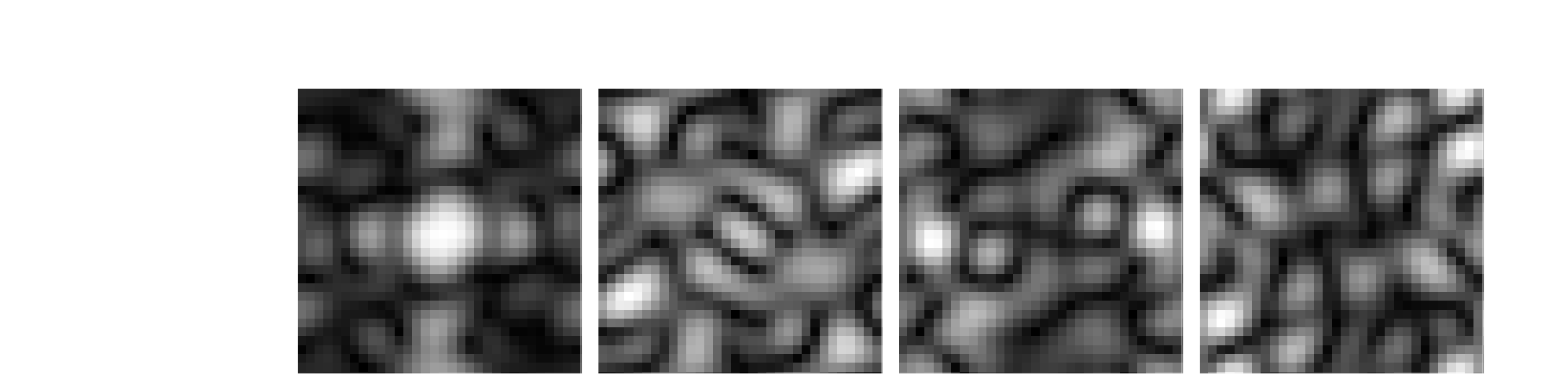}
  \label{fig:test3}
  \centering
  \includegraphics[trim = {12cm, 11cm, 15cm, 10cm}, clip,width=\textwidth]{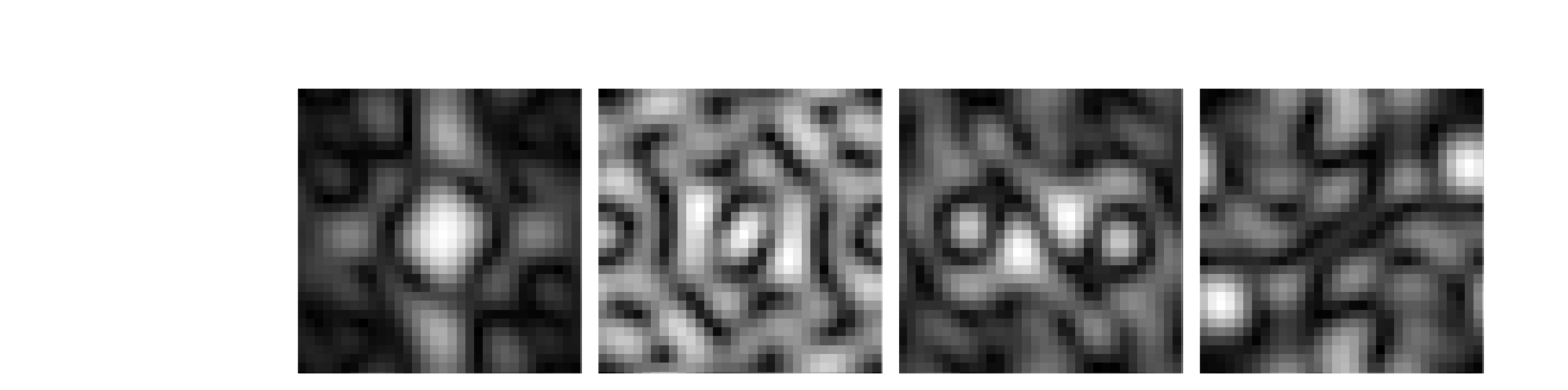}
  \subcaption{conv 6 filters}
\end{minipage}%
\vspace{-0.5in}
\caption{Visualizing filters of ReconNet \cite{lohit2017convolutional} in the frequency domain for MR = 0.25 and 0.10. We clearly see several similarities between the filters at the two MRs. This points to the possibility that the filters can be used across MRs.}
\label{fig:filter_vis}
\end{figure*}

\subsection{Motivation for three-stage training process} We will first describe the motivation through which the algorithm can be derived. We use the extended ReconNet architecture \cite{Kulkarni_2016_CVPR,lohit2017convolutional}, for insight into designing Rate-Adaptive CS. We first trained ReconNet at two different MRs = 0.25 and 0.1. Then, we visualized the filters in the convolutional layers. Although the filters are not exactly the same, one can immediately observe remarkable similarities in the filter structures. When observed in the frequency domain, both sets of filters have a ``speckle-field'' structure and more interestingly, even more similarities emerge. Figure \ref{fig:filter_vis} shows some pairs of filters, from every layer, at the two MRs that are very similar to each other. This may be because of the fact that output generated at the end of the second FC layer are image-like and have spatial correlated structure (observed by Lohit et al. \cite{lohit2017convolutional}) at both MRs, just differing in quality. Thus, similar convolutional operations are required to generate the final high-quality output. This leads to our idea that convolutional filters from one MR can be reused over the MR range of $[k,m]$. We later show empirically that this is also true for autoencoders, not just CNNs. 

Once the measurements are obtained from the spatial multiplexer (the first layers, $\Phi$, in the network), the second layer maps it back into a 2D array to obtain a pseudo-image. In our experiments, the second layer $\Psi$ is constrained to be $\Phi^\dagger$, the pseudoinverse of $\Phi$ given by $\Phi^T(\Phi\Phi^T)^{-1}$. This reduces the total number of parameters in the network by a large amount, especially for ReconNet-like architecture with convolutional layers, thus reducing the chance of overfitting. Now, we describe the three-stage training algorithm required to ensure that the network obeys the design constraints of Rate-Adaptive CS. As the convolutional filters are fixed for the entire range of MRs, the only difference in network architecture are the first ($\Phi$) and second ($\Psi$) FC layers. We will denote the parameters in the system, except $\Phi$ and $\Psi$, by $\Theta$. 

\vspace{5mm}

\noindent \textbf{Stage 1:} In the first stage, we train the convolutional layers (or the later FC layers in the case of autoencoders), for which we train the entire network at the base measurement rate of MR = $\frac{m}{n}$. That is, $\Phi$ is set to be of size $m \times n$. This ensures that the convolutional layers are most suited for the upper limit of the MR range, thus leading to the highest performance of the network at MR = $\frac{m}{n}$, as required. We call this the base network.

\vspace{5mm}

\noindent \textbf{Stage 2:} In the second stage, we freeze all the parameters $\Theta$. We set the size of $1^{\text{st}}$ FC layers to be of $k \times n$ (thus, size of $2^{\text{nd}}$ FC layer is $n \times k$), and optimize over only these parameters. This ensures that the network performs well at the MR = $\frac{k}{n}$.

\vspace{5mm}

\noindent \textbf{Stage 3:} In the third stage, we add a single row at a time to the $1^{\text{st}}$ FC layer and optimize over only the newly added variables. All the other variables, the remaining rows of $\Phi$, and $\Theta$, are held constant. \textbf{Thus, the output of the three-stage training algorithm is a measurement matrix $\Phi$ such that any subset of its rows $\Phi(1:r,:), k\leq r \leq m$, is a valid measurement matrix for the reconstruction/inference network defined by $\Psi(:,1:r)$ and $\Theta$.} The proposed training algorithm is summarized in Algorithm \ref{algo}. The general framework for jointly learning $\Phi$ and the reconstruction/inference network is illustrated in Figure \ref{fig:learning_mm}.

\subsection{Hardware Considerations}
Existing single-pixel camera (SPC) architectures allow for floating-point values in the learned $\Phi$. E.g., in Kerviche et al.[19], the values are converted to 9-bit signed integers and are normalized to the maximum value in $\Phi$. The mirrors in the DMD can either let light reach the sensor (ON) or reject it (OFF), but the fraction of measurement-time the mirror is in the ON state (duty-cycle) encodes the floating-point values in $\Phi$. It has also been demonstrated by \cite{Kulkarni_2016_CVPR,lohit2017convolutional} that very good image reconstruction using ReconNet is possible from this SPC architecture.

\begin{algorithm}\captionsetup{labelfont={sc,bf}, labelsep=newline}
\caption{Training algorithm for Rate-Adaptive CS.}
\begin{algorithmic}

    \REQUIRE {$k \longleftarrow$ Min. \#rows of $\Phi$, $m \longleftarrow$ Max. \#rows of $\Phi$, max\_iters\_1, max\_iters\_2, iters\_per\_row and other hyperparameters}
    \ENSURE {$\Phi_{new}$, $\Psi_{new} = \Phi_{new}^\dagger$ and $\Theta$, the remaining parameters in the reconstruction/inference network}
    
    \STATE $\text{Initialize}$ network with size of $1^{\text{st}}$ FC layer = size($\Phi) = m\times n$\\
   
   \textbf{\underline{Stage 1}}
    
    \For {iter = 1 \KwTo max\_iters\_1}{
    Optimize over $\Phi$ and $\Theta$\;
    }
       \textbf{\underline{Stage 2}}

    $\Phi_{k} \longleftarrow \Phi(1:k,:); \Psi_{k} = \Phi_{k}^\dagger$\;
    Replace $\Phi$ with $\Phi_{k}$ in the network\;

    \For {iter = 1 \KwTo max\_iters\_2}{ 
    Optimize over $\Phi_{k}$, holding $\Theta$ constant\;
    }
           \textbf{\underline{Stage 3}}

    $\Phi_{new} \longleftarrow \Phi_{k}; \Psi_{new} = \Phi_{new}^\dagger$\;
    \For {$r = k+1$ \KwTo $m$}{   
    	\For {iter = 1 \KwTo iters\_per\_row}{
        $\Phi_{new} \longleftarrow [\Phi_{new}; \Phi(r,:)]; \Psi_{new} = \Phi_{new}^\dagger$\; 
        Optimize over $\Phi(r,:)$, holding $\Phi(1:r-1,:)$ and $\Theta$ constant\;
        }
    }

    \label{algo}
    
\end{algorithmic}
\end{algorithm}

\subsection{Comparison with a random Gaussian $\Phi$}
Earlier works in literature such as Mousavi et al. [28] and Lohit et al. [24] have convincingly demonstrated that for the same network architecture, learning $\Phi$ jointly with the reconstruction algorithm results in substantially improved ($\mathtt{\sim}3$dB) PSNRs over random $\Phi$. In our problem, we have made similar observations: that using a random $\Phi$ performs worse than learning the $\Phi$, especially when one starts dropping rows. We did not include this in the paper, as it is now commonly known.

\section{Experimental results on Rate-Adaptive CS} \label{sec:Rate-Adaptive CS_expt}
In this section, we describe experimental results for the Rate-Adaptive CS framework (Section \ref{sec:Rate-Adaptive CS}, Algorithm \ref{algo}), where we learn $\Phi$ and the reconstruction/inference n/w that can be operated over a range of MRs. We compare our results with the \textbf{``vanilla'' framework} \cite{lohit2017convolutional,adler2017block}, where the system is trained for a single MR. We carry out experiments for both image reconstruction as well image recognition. As mentioned, we need to input the values of the operating MR range $[k,m]$ to the algorithm, which are the minimum and maximum values of MR for which the system can operate. We demonstrate that our algorithm is effective as follows. Given a trained network ($\Phi, \Psi, \Theta$), we start with $\Phi(1,:)$ as the measurement matrix and $\Psi(:,1)$ as the second FC layer. We observe the performance of the system on the test set. Then, we add one row at a time to the measurement matrix, and one column at a time to the second FC layer and measure the change in performance, compared to the vanilla algorithm which is trained for a single MR.

\subsection{Rate-Adaptive CS for image reconstruction}
In this section, we will describe the network, the training and testing protocols used for the image reconstruction problem: given $\mathbf{y}$, return $\mathbf{x}$, where $\mathbf{y} = \Phi\mathbf{x}$. It is important to note that $\Phi$ is also learned (see Figure \ref{fig:learning_mm}), and forms the first layer of the network while training. We experiment on two network architectures -- \textbf{the extended ReconNet architecture \cite{lohit2017convolutional} which uses 6 convolutional layers, and a 3-layer autoencoder \cite{mousavi2015deep}} -- for jointly learning the measurement operator and the reconstruction algorithm. Note that the networks are designed for block-wise reconstruction which means that the sensing and reconstruction process is for non-overlapping blocks of the image rather than the entire image. The loss function is simply the Euclidean loss between the desired output $\mathbf{x}$ (the original image itself) and the estimated reconstruction $\mathbf{\hat{x}}$.

\subsubsection{Datasets}

\begin{figure*}[!htb]
	\centering
	\begin{subfigure}[]{0.15\textwidth}
		\centering
		\includegraphics[height=0.8in]{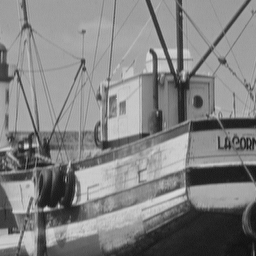}
		\caption{Boats}
	\end{subfigure}%
	\hfill
	\begin{subfigure}[]{0.15\textwidth}
		\centering
		\includegraphics[height=0.8in]{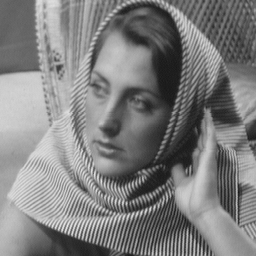}
		\caption{Barbara}
	\end{subfigure}%
	\hfill
	\begin{subfigure}[]{0.15\textwidth}
		\centering
		\includegraphics[height=0.8in]{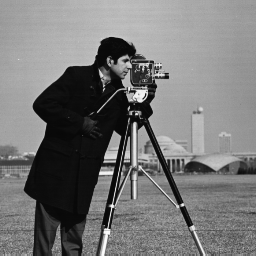}
		\caption{\tiny Cameraman}
	\end{subfigure}%
	\hfill
	\begin{subfigure}[]{0.15\textwidth}
		\centering
		\includegraphics[height=0.8in]{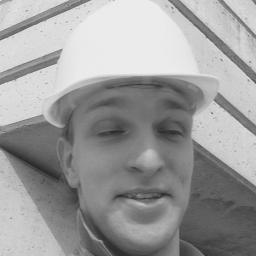}
		\caption{Foreman}
	\end{subfigure}%
	\hfill
	\begin{subfigure}[]{0.15\textwidth}
		\centering
		\includegraphics[height=0.8in]{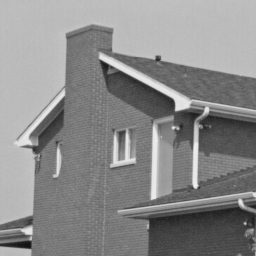}
		\caption{House}
	\end{subfigure}%
	\hfill
	\begin{subfigure}[]{0.15\textwidth}
		\centering
		\includegraphics[height=0.8in]{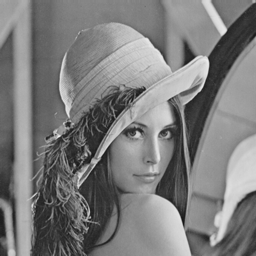}
		\caption{Lena}
	\end{subfigure}%
	
	\hfill
	\begin{subfigure}[]{0.18\textwidth}
		\centering
		\includegraphics[height=0.8in]{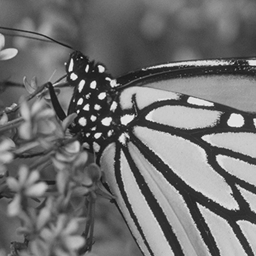}
		\caption{Monarch}
	\end{subfigure}
	\hfill
	\begin{subfigure}[]{0.18\textwidth}
		\centering
		\includegraphics[height=0.8in]{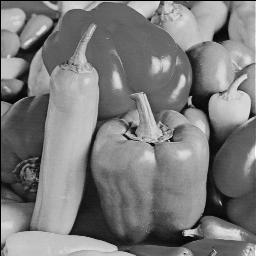}
		\caption{Peppers}
	\end{subfigure}
	\hfill
	\begin{subfigure}[]{0.18\textwidth}
		\centering
		\includegraphics[height=0.8in]{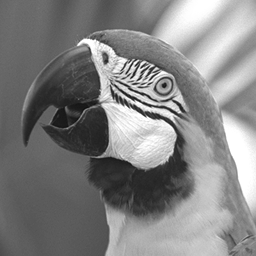}
		\caption{Parrots}
	\end{subfigure}
	\hfill
	\begin{subfigure}[]{0.18\textwidth}
		\centering
		\includegraphics[height=0.8in]{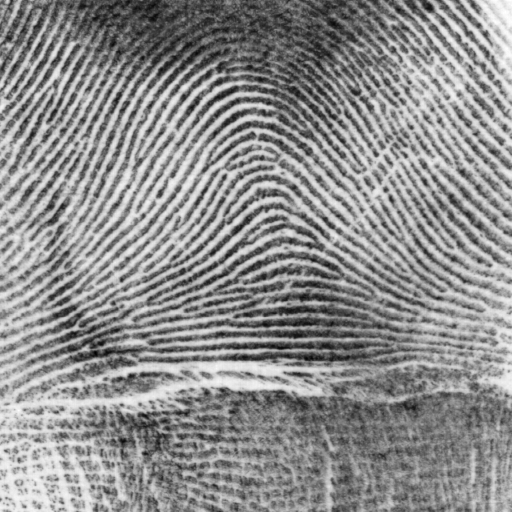}
		\caption{Fingerprint}
	\end{subfigure}
	\hfill
	\begin{subfigure}[]{0.18\textwidth}
		\centering
		\includegraphics[height=0.8in]{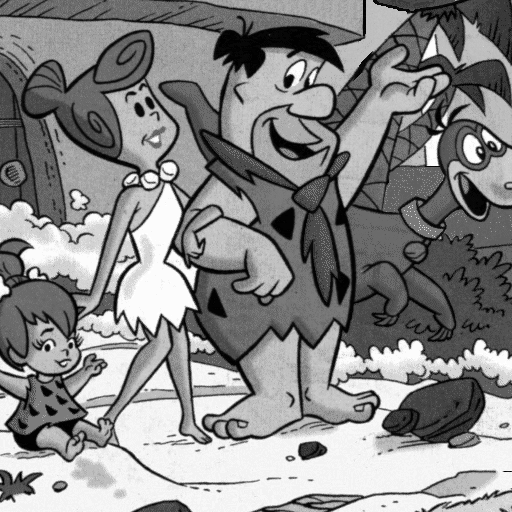}
		\caption{Flintstones}
	\end{subfigure}%
	
	\caption{Test set images used for evaluating the algorithms for the image reconstruction application. Note that all images are of size $256 \times 256$ (~64 non-overlapping $33 \times 33$ blocks) except Fingerprint and Flintstones which are $512 \times 512$ (~256 non-overlapping $33 \times 33$ blocks)} 
	\label{fig:test_images}
    \vspace{-0.2in}
\end{figure*}

The training and test sets are identical to those employed in \cite{Kulkarni_2016_CVPR}. The training set contains 91 natural images that can be downloaded from this website \footnote{\url{mmlab.ie.cuhk.edu.hk/projects/SRCNN/SRCNN_train.zip}}. The training set for the network is obtained by constructing image blocks of size $33 \times 33$. The test set, shown in Figure \ref{fig:test_images} consists of 11 standard images employed in image processing literature obtained from these two links given here \footnote{\url{https://web.archive.org/web/20160403234531/http://dsp.rice.edu/software/DAMP-toolbox}} \footnote{\url{http://see.xidian.edu.cn/faculty/wsdong/NLR\textunderscore Exps.htm}}.

\subsubsection{Network architectures}
We use two network architectures in our image reconstruction experiments. They are described below. 

\vspace{5mm}

\noindent\textbf{ReconNet:} We use the network architecture presented in \cite{lohit2017convolutional} for jointly learning the measurement matrix as well as the image reconstruction algorithm. The first layer is a fully connected (FC) layer of size $m \times n$ which serves as the measurement operator $\Phi$. The second layer is also an FC layer of size $n \times m$ which converts the output of the first layer into the same dimension as the original signal and is then reshaped to form a 2D array of the same dimensions as the image. This is then followed by two ReconNet units. Each ReconNet unit consists of 3 convolutional layers with ReLU activation, except for the last layer of the second ReconNet unit. The first conv layer produces 64 feature maps with a $11 \times 11$ filters. The second conv layer converts this to 32 feature maps with $1 \times 1$ filters. The third conv layer produces a single feature map with $7 \times 7$ filters. The output of the last layer is the reconstructed image block. At test time, the CS measurements pass through the second FC layer and 6 conv layers to yield the reconstructed image block. 

\vspace{5mm}

\noindent\textbf{Autoencoder}:
The network architecture is the same as the one used by Mousavi et al. \cite{mousavi2015deep}. The first layer is an FC layer of size $m \times n$ which serves as the measurement operator $\Phi$. The second layer is also an FC layer of size $n \times m$ which converts the output of the first layer into the same dimension as the original signal and is then reshaped to form a 2D array of the same dimensions as the image block. This is followed by another two FC layers of sizes $m \times n$ and $n \times m$ respectively. The output of the last layer is the reconstructed image block. We use ReLU non-linearities in all the intermediate FC layers. At test time, the CS measurements obtained from the compressive imager/spatial multiplexer is passed through second, third and fourth layers (3 layers) to get the reconstruction.

\subsubsection{Training details} For the vanilla algorithm, we train all networks for $5 \times 10^5$ iterations with the Adam optimizer \cite{kingma2015adam} with an initial learning rate of $10^{-4}$. The best model is chosen based on a validation set. For the rate-adaptive framework, the training algorithm is shown in Algorithm \ref{algo} and optimization of individual stages is performed using Adam optimizer with a learning rate of $10^{-4}$ and max\_iters\_1 $= 3\times10^5$, max\_iters\_2 $= 2\times10^5$ and iters\_per\_row $= 500$. These are chosen so as to keep the number of iterations approximately same as that of the vanilla training algorithm, $5\times10^5$ iterations. This allows for easier and fair comparison. However, we note that these are hyperparameters and further tuning could provide improved results.

\subsubsection{Results} 
\begin{figure*}
\centering
\includegraphics[width=0.8\textwidth]{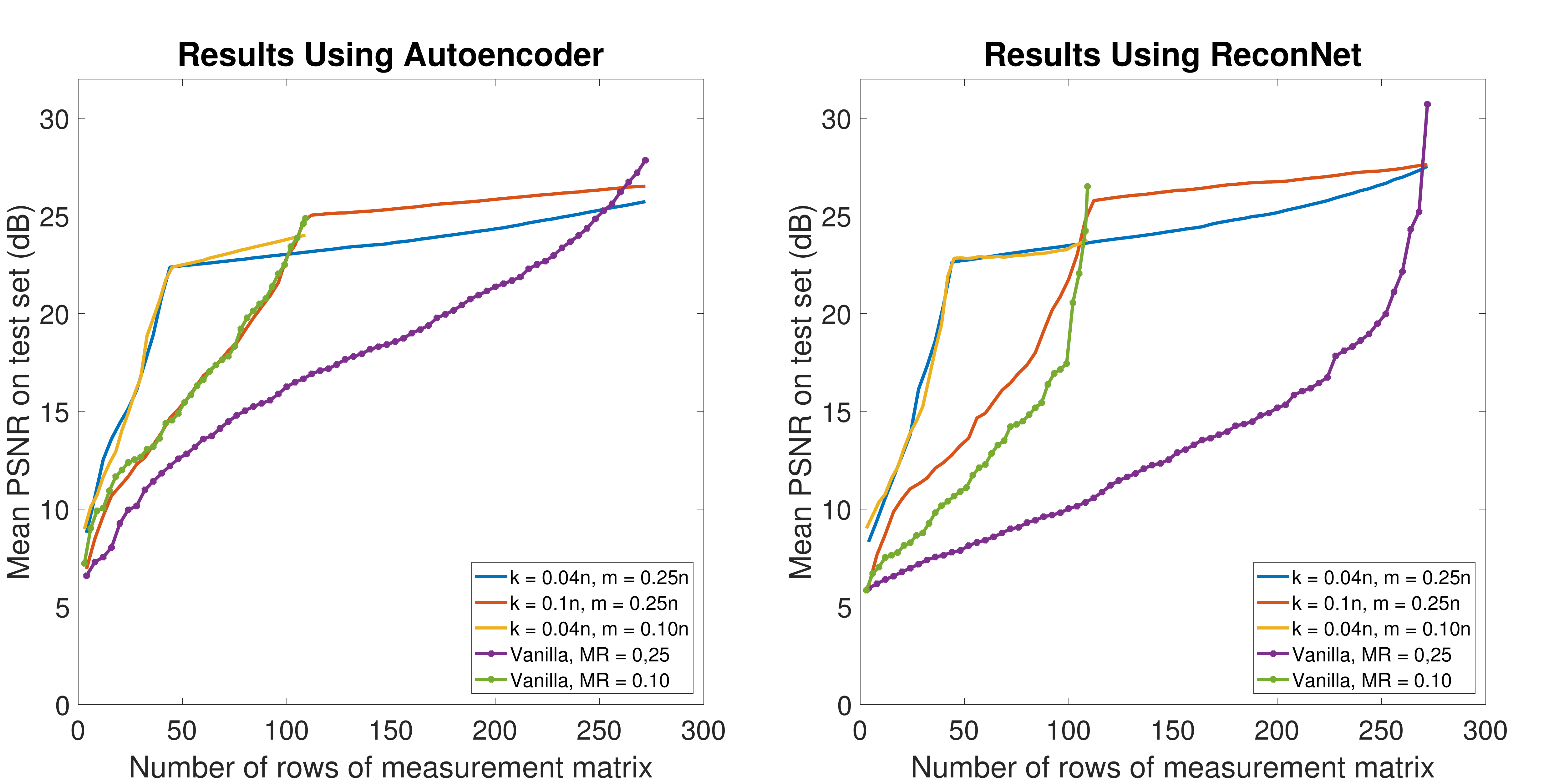}
\caption{Rate-Adaptive CS results for image reconstruction for the Rate-Adaptive CS framework compared to the vanilla training algorithm trained for a single measurement rate. We test on two architectures -- autoencoder \cite{mousavi2015deep} and ReconNet \cite{lohit2017convolutional}. Clearly, Rate-Adaptive CS is stable over the entire chosen operating range while the performance of the vanilla framework drops considerably in the same range. Best viewed in color.}
\label{fig:rate_adaptive_cs_plots_reconstruction}
\end{figure*}

\begin{figure*}
\centering
\begin{subfigure}[b]{\textwidth}
   \includegraphics[scale=0.35,trim={2cm 9.5cm 1.7cm 7.5cm},clip,width=\textwidth]{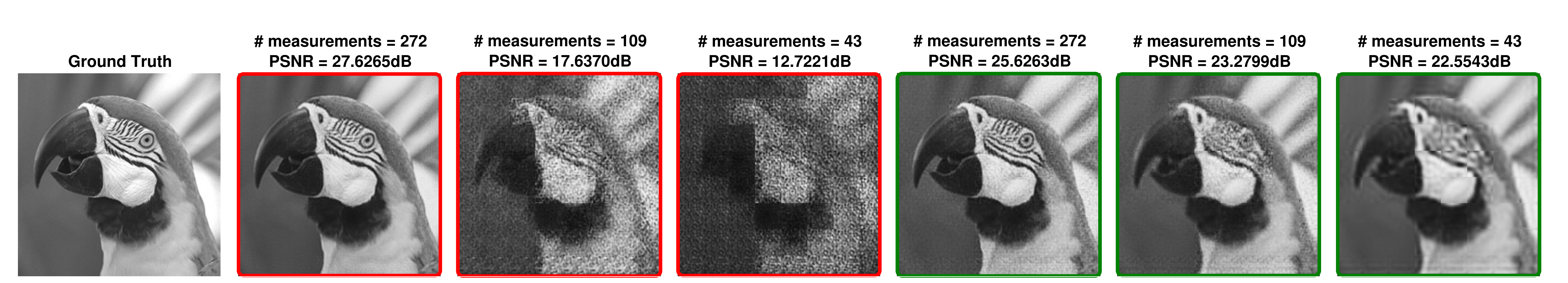}
\end{subfigure}
\begin{subfigure}[b]{\textwidth}
   \includegraphics[scale=0.35,trim={2cm 9.5cm 1.7cm 7.5cm},clip,width=\textwidth]{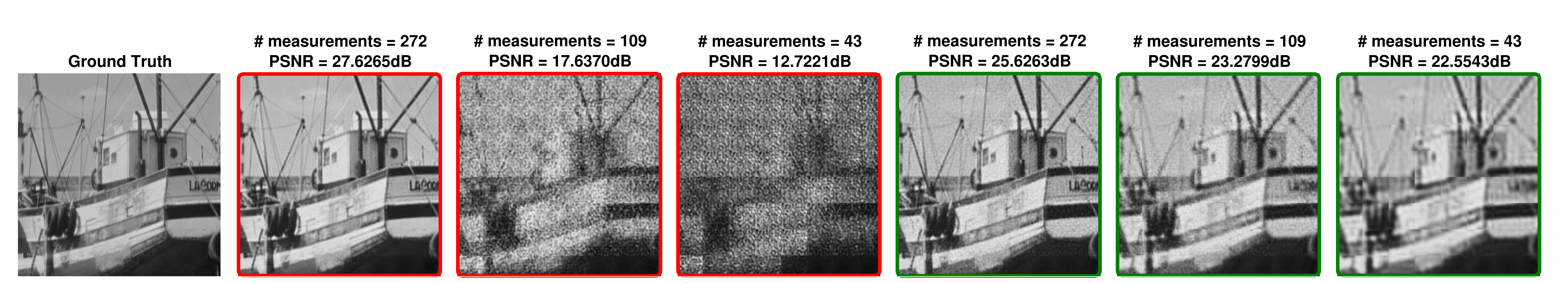}
\end{subfigure}
\caption{The figures show examples of reconstruction for 2 test set images using the vanilla training algorithm \cite{lohit2017convolutional} trained for MR = 0.25, compared to Rate-Adaptive CS, proposed in this paper ($[k,m] = [0.04n,0.25n])$ We observe that Rate-Adaptive CS performs significantly better than the vanilla approach over a range of MRs. An autoencoder\cite{mousavi2015deep} is used as the reconstruction network ($\Theta$).}
\label{fig:Rate-Adaptive_CS_vis_autoencoder}
\end{figure*}

\begin{figure*}
\centering
\begin{subfigure}[b]{\textwidth}
   \includegraphics[scale=0.35,trim={2cm 9.5cm 1.7cm 7.5cm},clip,width=\textwidth]{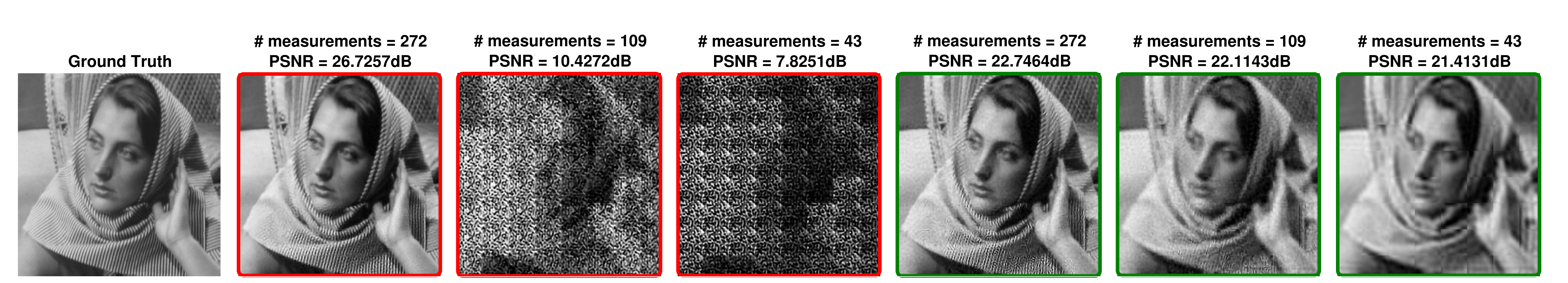}
\end{subfigure}
\begin{subfigure}[b]{\textwidth}
   \includegraphics[scale=0.35,trim={2cm 9.5cm 1.7cm 7.5cm},clip,width=\textwidth]{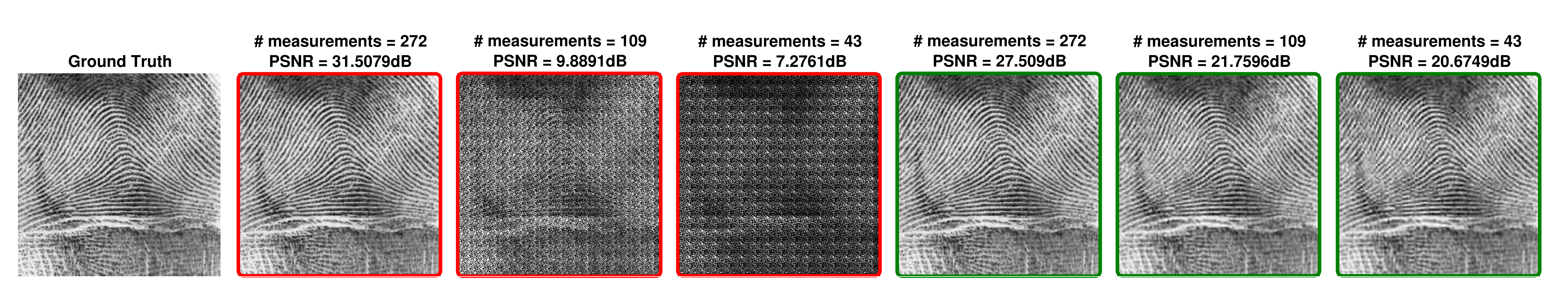}
\end{subfigure}
\begin{subfigure}[b]{\textwidth}
   \includegraphics[scale=0.35,trim={2cm 8cm 1.7cm 7.5cm},clip,width=\textwidth]{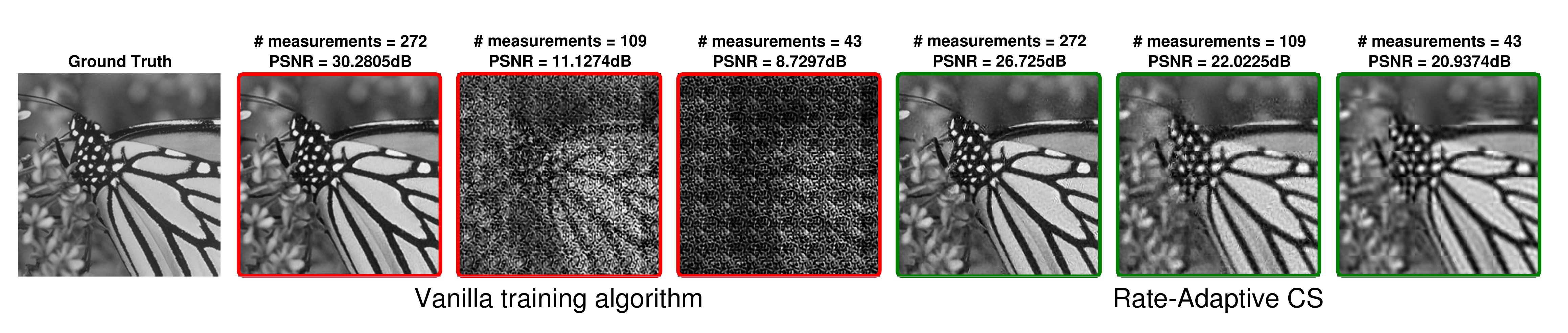}
\end{subfigure}
\caption{The figures show examples of reconstruction for 3 test set images using the vanilla training algorithm \cite{lohit2017convolutional} trained for MR = 0.25, compared to Rate-Adaptive CS, proposed in this paper ($[k,m] = [0.04n,0.25n])$ We observe that Rate-Adaptive CS performs significantly better than the vanilla approach over a range of MRs. ReconNet\cite{lohit2017convolutional} is used as the reconstruction network ($\Theta$).}
\label{fig:Rate-Adaptive_CS_vis_reconnet}
\end{figure*}

Figure \ref{fig:rate_adaptive_cs_plots_reconstruction} and Table \ref{table:stable_reconnet} show the results in terms of the mean PSNR obtained on the test set for different combinations of $[k,m]$. Figures \ref{fig:Rate-Adaptive_CS_vis_autoencoder} and \ref{fig:Rate-Adaptive_CS_vis_reconnet} shows the reconstructions for a subset of test images comparing the vanilla framework with Rate-Adaptive CS with the autoencoder and ReconNet respectively. 


We can clearly observe the desired behavior of Rate-Adaptive CS. For all cases tested, the performance only decreases gradually for Rate-Adaptive CS, as the number of measurements is decreased, within the operating range. This is not true in the case of the vanilla algorithm, where the performance falls much more quickly when the test MR strays from the training MR. It also appears that the performance at a measurement rate depends on the value of $k$. For example, a lower $k = 0.04n$ leads to a lower mean PSNR at MR = 0.10. This can be explained by observing that Stage 3 of the algorithm builds on the $\Phi$ learned in Stages 1 and 2 and is expected to be sub-optimal compared to the vanilla training algorithm for the specific MR=0.10. Naturally, the performance at MR = $\frac{m}{n}$ depends on $m-k$, but only to a small extent, as required. This can be observed easily from the plots. Using ReconNet as the underlying architecture, we note that a rate-adaptive network for $MR = [0.10,0.25]$ performs on average 9 db and up to 15.2 dB better than a vanilla network trained for $MR = 0.25$, when tested over all MRs. Similar results are observed for all other cases. We note in passing that these results can be further improved by using an adversarial loss term in addition to the Euclidean loss.

\begin{table*}
\normalsize
\centering

\begin{tabular}{|c|c|p{0.8cm}|p{0.8cm}|p{0.8cm}|p{0.8cm}|p{0.8cm}|p{0.8cm}|}
\hline
\multirow{2}{*}{Method} & \multirow{2}{*}{\makecell{Training MR \\ (range)}} & \multicolumn{3}{c|}{\makecell{Autoencoder\\Test MR}} & \multicolumn{3}{c|}{\makecell{ReconNet\\Test MR}}\\
\cline{3-8}
& & 0.25 & 0.10 & 0.04 & 0.25 & 0.10 & 0.04 \\
\hline
Vanilla \cite{mousavi2015deep,lohit2017convolutional} & 0.25 & 27.85 & 16.83 & 12.19 & 30.72 & 10.35 & 7.74 \\
\hline
Vanilla & 0.10 & N/A & 24.88 & 14.45 & N/A & 26.51 & 10.51 \\
\hline
Vanilla & 0.04 & N/A & N/A & 22.20 & N/A & N/A & 22.49 \\
\hline
Rate-Adaptive CS & $0.04 - 0.25$ & 25.73 & 23.14 & 22.37 & 27.54 & 23.62 & 22.62 \\
\hline
Rate-Adaptive CS & $0.10 - 0.25$ & 26.52 & 25.02 & 14.45 & 27.62 & 25.73 & 12.61 \\
\hline
Rate-Adaptive CS & $0.04 - 0.10$ & N/A & 24.00 &22.33 & N/A & 23.76 & 22.79 \\
\hline
\end{tabular}
\caption{Rate-Adaptive CS (Algorithm \ref{algo}) versus vanilla training algorithm for the image reconstruction problem in terms of mean PSNR (dB) on the test set. The vanilla algorithm trains the network for a single MR, while Rate-Adaptive CS is trained so that, at test time, the system is stable over the entire chosen MR range. This can be easily observed from the results shown. $n=1089$.}
\label{table:stable_reconnet}
\end{table*}

\subsubsection{How rate-adaptation modifies measurement operators} 
\begin{figure*}[ht!]
\begin{minipage}[c][5cm][t]{.33\textwidth}
  \centering
  \includegraphics[trim = {6cm, 12cm, 6cm, 8cm}, clip,width=\textwidth]{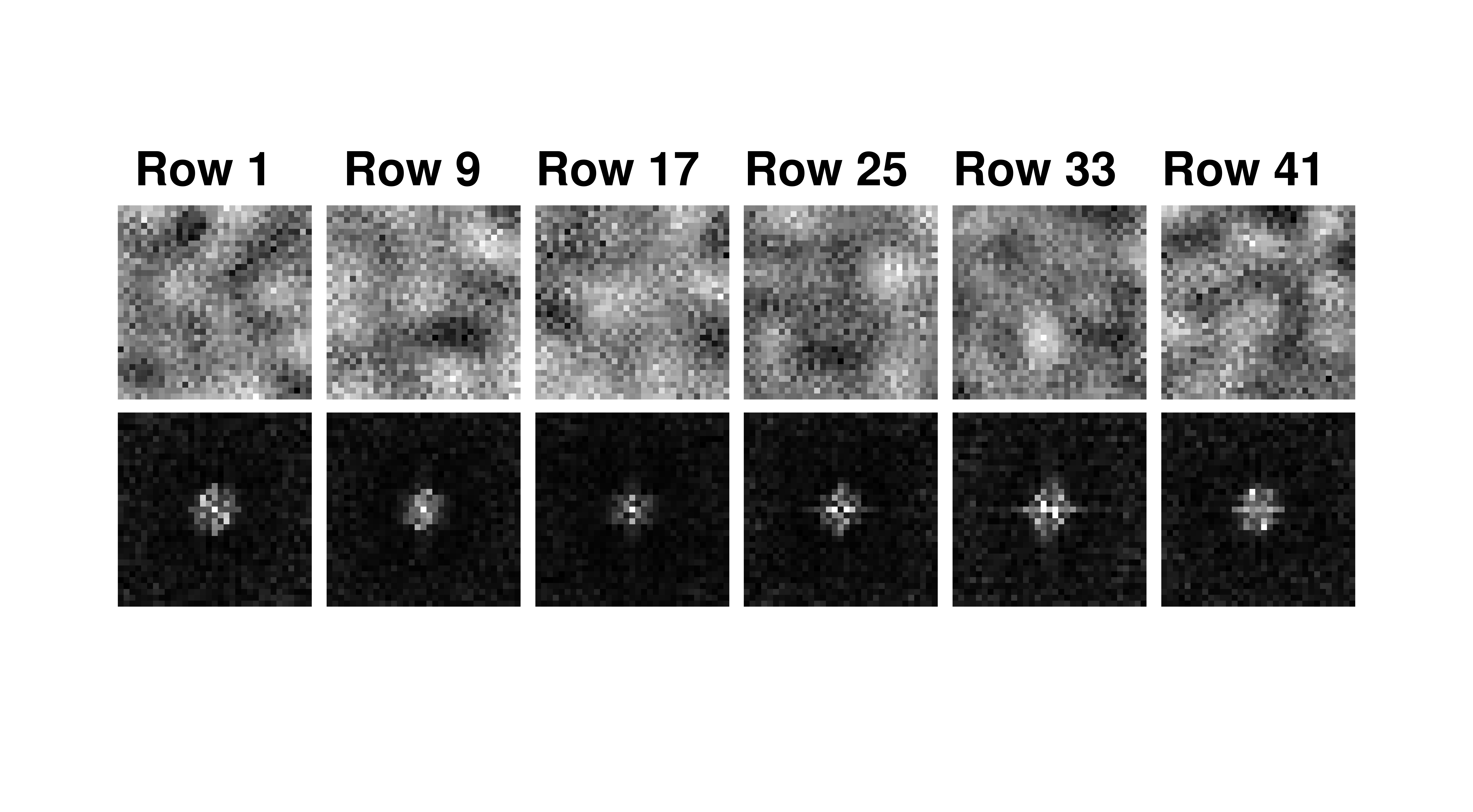}
  \subcaption{Vanilla $\Phi$, MR = 0.04}
\end{minipage}%
\begin{minipage}[c][5cm][t]{.33\textwidth}
  \centering
  \includegraphics[trim = {6cm, 12cm, 6cm, 8cm}, clip,width=\textwidth]{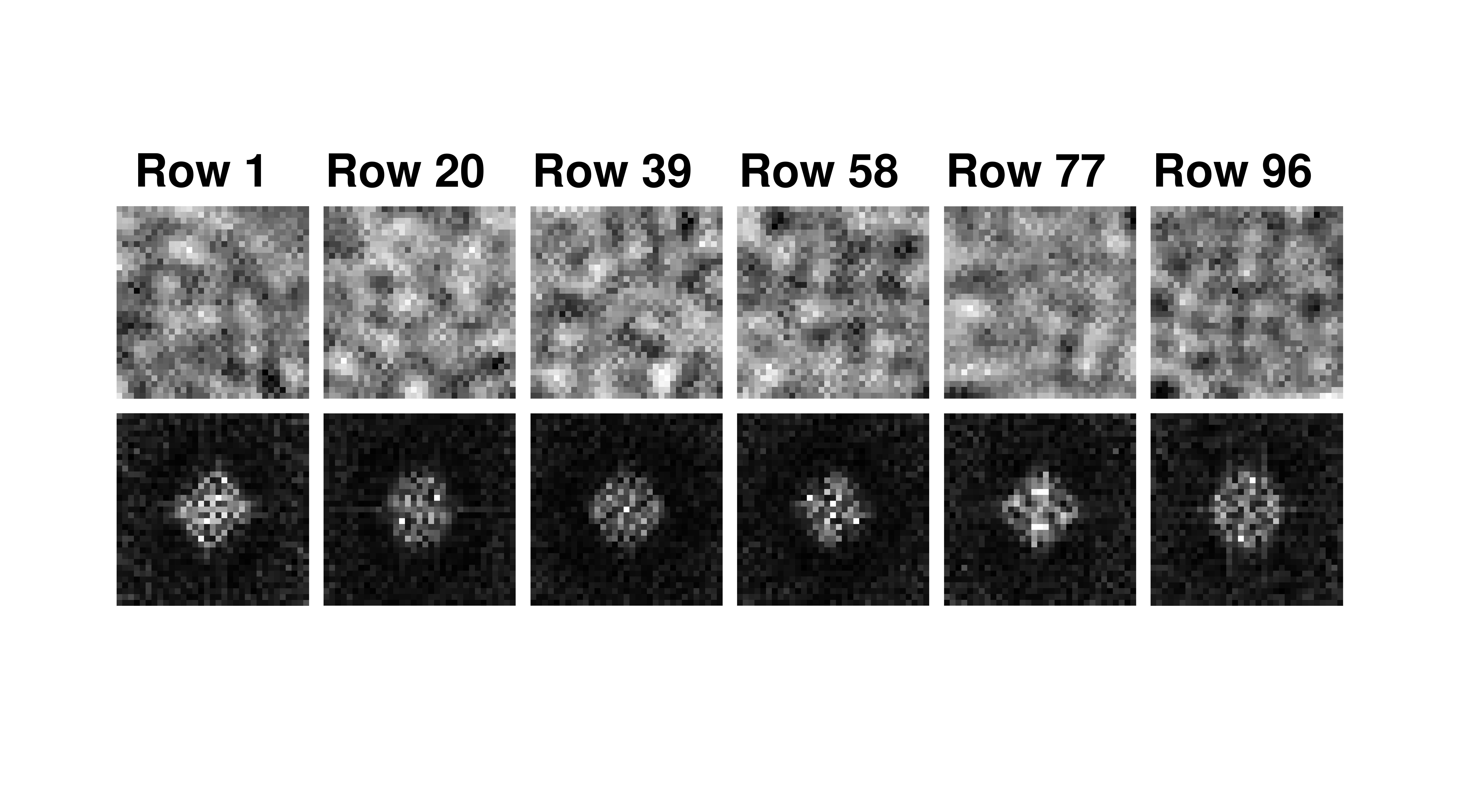}
    \subcaption{Vanilla $\Phi$, MR = 0.10}
\end{minipage}%
\begin{minipage}[c][5cm][t]{.33\textwidth}
  \centering
  \includegraphics[trim = {6cm, 12cm, 6cm, 8cm}, clip,width=\textwidth]{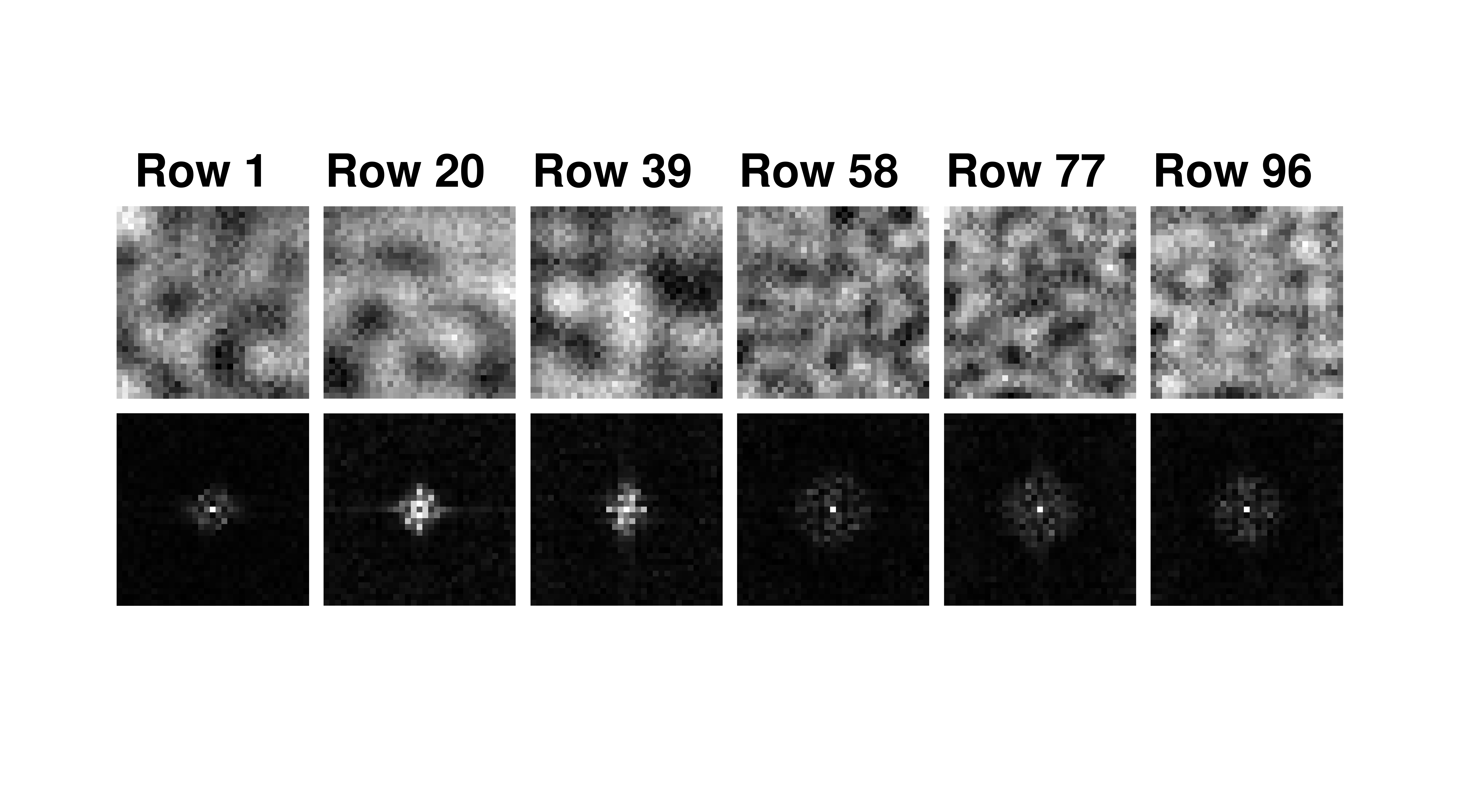}
  \subcaption{Rate-Adaptive $\Phi$}
\end{minipage}%
\vspace{-0.7in}
\caption{The figures show the visualization of the rows of the learned $\Phi$ in the spatial (top) and Fourier (bottom) domains for the ReconNet architecture. The earlier rows of $\Phi$ in the case of Rate-Adaptive CS for the MR range [0.04,0.10] resemble the rows of $\Phi$ obtained using the vanilla training algorithm at MR = 0.04, while the later rows look similar to the rows of the vanilla $\Phi$ for MR = 0.10.}
\label{fig:Phi_vis}
\end{figure*}

Here, we compare the $\Phi$ learned in the Rate-Adaptive framework, with the vanilla algorithm in Figure \ref{fig:Phi_vis}. We can observe that the rows of rate-adaptive $\Phi$ look like a sampling of rows of $\Phi$ trained at different MRs. For instance, rows $1, 20, 39$ in column (c) are visually similar to the images shown in column (a), whereas rows $58, 77, 96$ in column (c) are similar to the ones in column (b). These observations suggest that rate-adaptive $\Phi$ shares characteristics of vanilla $\Phi$'s across a range of MRs.

\subsection{Rate-Adaptive CS Reconstruction for Object Tracking}
We use object tracking in order to provide a proof of concept for Rate-Adaptive CS. In our framework, the frames of the video for object tracking are reconstructed using the measurements acquired by the spatial multiplexer that is learned in conjunction with the reconstruction algorithm, and the goal is to track the main object in the scene. We choose ReconNet \cite{lohit2017convolutional} as the underlying network architecture. We use Kernelized Correlation Filter (KCF) \cite{henriques2015high}, an off-the-shelf tracker, for our experiments. We use HoG features for the images and Gaussian kernel for the tracker. We use a well-known dataset of 15 publicly available videos \cite{wu2015object} \footnote{BlurBody, BlurCar1, BlurCar2, BlurCar4, BlurFace,
BlurOwl, Car2, CarDark, Dancer, Dancer2, Dudek,
FaceOcc1, FaceOcc2, FleetFace, Girl2}. We consider three different simple MR adaptation schemes in order to illustrate how Rate-Adaptive CS can be used for sample-efficient object tracking:

(a) Linearly decreasing MR: For a given video, we use an initial MR$=m$, and steadily decrease such that the final frame of the video is acquired at MR$=k$. We compare the performance of Rate-Adaptive CS with that of the vanilla framework, for the case where the MR is decreased linearly with the frame number. In this experiment, we choose the operating MR range to be $[k,m]=[0.04,0.25]$, and the vanilla network is trained at MR$=0.25$. This adaptation scheme could be utilized for an energy/memory constrained application. The initial MR is set to $m=0.25$.

(b) Content-based MR adaptation using Euclidean loss: Here, we compute the normalized Euclidean loss between successive frames of the video. If the difference is smaller $\alpha$, we reduce the number of measurements for the next frame by a fixed amount $\Delta MR$. If the difference is larger than $\beta$, we increase the number of measurements for the next frame by $\Delta MR$. This scheme can be viewed as a simple form of motion-based MR adaptation. For the experiment, we choose the operating MR range to be $[0.04, 0.25]$, and the vanilla network is trained at MR$ = 0.25$, $\alpha = 0.15, \beta = 0.3, \Delta MR = 3$.

(c) Content-based MR adaptation using maximum correlation: For each frame of the video, the tracking algorithm outputs the maximum response value of cross-correlation $\in [0,1]$ between the templates and the frame, which indicates the confidence of the tracker, and is used to localize the object. We utilize this value to determine the MR for sensing the next frame. If the max. correlation is smaller than $\gamma$, we increase the number of measurements for the next frame by a fixed amount $\Delta MR$. If the max. correlation is larger than $\gamma$, we decrease the number of measurements for the next frame by $\Delta MR$. In this experiment, we choose the operating MR range to be $[0.04, 0.10]$ , and the vanilla network is trained at MR$ = 0.25$, $\gamma=0.3, \Delta MR = 3$.

In each of the above cases, we compare the performance of the reconstructions obtained using Rate-Adaptive CS to the vanilla training algorithm, using ReconNet as the reconstruction architecture. The performance is measured in terms of the average precision of the localizations, for a pixel error threshold of 20 pixels. In each case, we also calculate the average MR over the entire database, that each of the heuristics lead to.

\begin{figure*}[!htb]
\centering
\includegraphics[trim = {1.8cm, 10cm, 2.6cm, 10cm} ,clip,width=\textwidth]{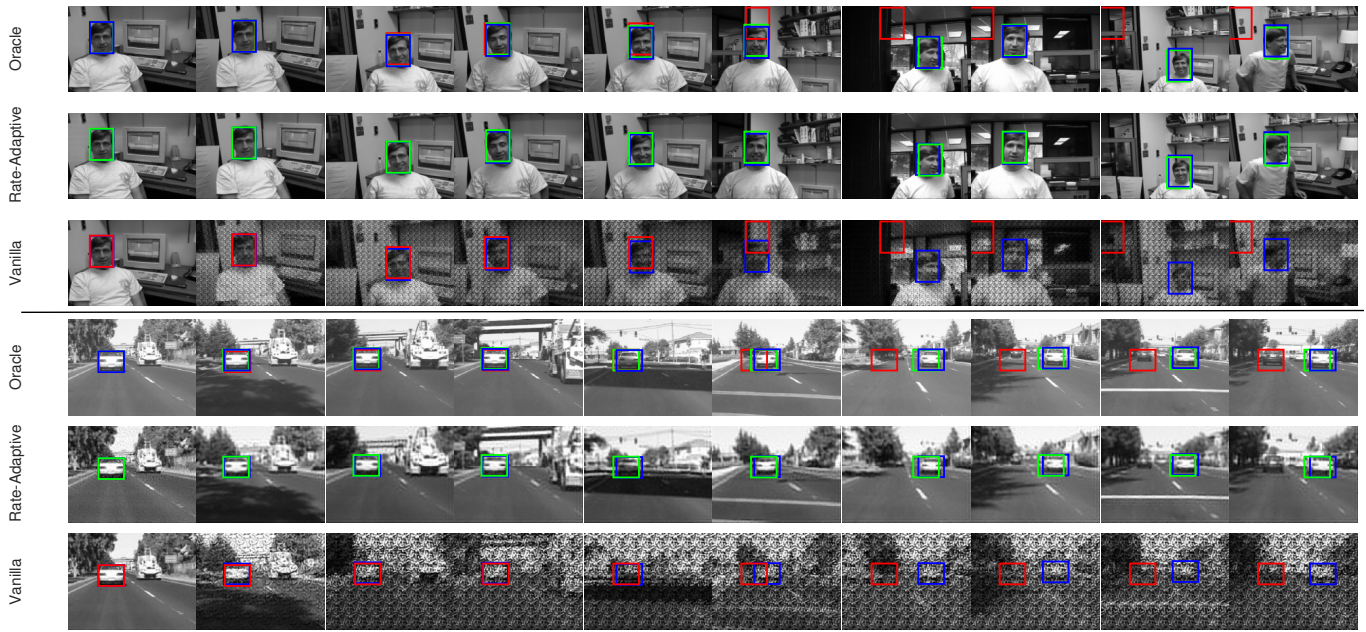}
\caption{Visual results on two videos for the object tracking, using maximum correlation of the detector to dynamically vary MR. For each video, the top row shows the frames acquired with no compression (conventional imaging, referred to as the oracle). The second and third rows display the reconstructions using Rate-Adaptive ReconNet (trained for MR = $[0.04, 0.10]$ and vanilla ReconNet (trained for MR = 0.10) respectively. \textcolor{blue}{Blue}, \textcolor{green}{green} and \textcolor{red}{red} boxes show the object locations for ground-truth, Rate-Adaptive ReconNet and vanilla ReconNet respectively. Unlike the rate-adaptive framework, as MR is varied, the reconstructions are very poor in the vanilla case, leading to poor tracking performance.}
\label{fig:tracking_vis}
\end{figure*}

\noindent \textbf{Results} The results are shown in Table \ref{table:tracking} and visualizations are provided in Figure \ref{fig:tracking_vis}. Clearly, as expected, Rate-Adaptive ReconNet is superior in terms of the tracking performance by a huge margin, because the vanilla network yields poor reconstructions at MRs for which it is not trained specifically. For comparison, the tracking performance with full images (oracle, i.e no compression) is about 80\% for a 20 pixel error threshold. Also, for the heuristics and the thresholds chosen, the average MR for the Rate-Adaptive case is much lower, and at the same time our approach maintains high tracking performance. We also observe that the MR determination based on maximum correlation is superior to the one based on Euclidean loss between successive frames. It is possible to have more sophisticated approaches to determining the best MR for each frame, and optimize the values of $\alpha, \beta, \gamma$ and $\Delta MR$, but such an elaborate study is beyond the scope of this paper. 

\begin{table*}
\normalsize
\renewcommand{\tabcolsep}{1.5mm}
\centering
\begin{tabular}{|c|c|c|c|c|c|c|}
\hline
\multirow{3}{*}{Method} & \multicolumn{6}{c|}{MR adaptation approach} \\
\cline{2-7}
& \multicolumn{2}{c|}{Linear Decrease} & \multicolumn{2}{c|}{Euclidean Loss} & \multicolumn{2}{c|}{Detector Confidence}\\
\cline{2-7}
& \makecell{Average \\ Precision} & Avg. MR & \makecell{Average\\ Precision} & Avg. MR & \makecell{Average \\ Precision} & Avg. MR \\
\hline
Vanilla & 54.29 \% & 0.1444 & 46.08 \% & 0.1113 & 63.31 \% & 0.0749\\
\hline
\makecell{Rate-Adaptive \\ CS} & 79.56 \% & 0.1444 & 70.82 \% & 0.1106 & 77.47 \% & 0.0462\\
\hline
\end{tabular}
\caption{Object tracking performance comparison of the Rate-Adaptive framework with the single-MR vanilla framework for dynamically varying MR. Results clearly show that Rate-Adaptive CS is superior in terms of average precision as well as the average number of measurements made.}
\label{table:tracking}
\end{table*}

\subsection{Rate-Adaptive CS for Image Recognition}

In this section, we extend the ideas presented for the image reconstruction problem to a very different task of image recognition, in order to demonstrate the wide applicability of the proposed method. We would like to perform image/object recognition directly on the multiplexed measurements obtained from the camera, bypassing reconstruction. Also note that the measurement operator learned in this case is optimized for the task of image recognition. We use two widely used datasets -- MNIST and CIFAR-10 -- for this purpose.

\subsubsection{Datasets}
As mentioned in the main paper we test our Rate-Adaptive CS algorithm on image recognition task from CS measurements directly, bypassing the reconstruction. For this task we use MNIST \cite{lecun1998gradient} and CIFAR-10 datasets \cite{krizhevsky2009learning}. The MNIST dataset consists of 50000 training and 10000 testing grayscale images of hand-written digits 0-9 of size $28 \times 28$. Thus $n = 784 (28 \times 28)$. CIFAR-10 dataset consists of $32 \times 32$ color images belonging to 10 classes with 50000 training images and 10000 testing images.

\subsubsection{Network architectures and training details}
\textbf{MNIST hand-written digit recognition:}
The network is a modified version of the LeNet-5 architecture\cite{lecun1998gradient} with 3 convolutional layers, two FC layers and a softmax layer. We add two FC layers at the input of the network, of which the first serves as $\Phi$ and the second FC layer is the matrix $\Psi$. The loss function is the cross-entropy loss between the estimated and the desired distribution. For the rate-adaptive framework, the optimization of the individual stages is performed using Adam optimizer with an initial learning rate of $10^{-4}$ and max\_iters\_1 = max\_iters\_2 $= 5\times10^5$ and iters\_per\_row $= 1000$. The vanilla training algorithm stops after Stage 1 of the rate-adaptive framework.

\noindent\textbf{CIFAR-10 Image Recognition:}
We modify the 32 layer resnet model for CIFAR \cite{he2016deep} by adding two fully connected layers at the input of network similar to MNIST model. We use cross entropy as loss function and momentum optimizer to optimize individual stages with max\_iters\_1 $= 2\times10^5$, max\_iters\_2 $= 5\times10^4$ and iters\_per\_row $= 1000$. We use a weight decay of $2\times10^{-4}$ and momentum of $9\times10^{-1}$. In the rate-adaptive framework, for Stage 1 we start with initial learning rate of $1\times10^{-1}$ and divide it by 10 at 40k, 80k and 120k iterations. For Stages 2 and 3 we use fixed learning rate of $1\times10^{-4}$. The vanilla training algorithm stops after Stage 1 of the rate-adaptive framework.

\begin{figure*}[!htb]
\centering
\includegraphics[trim = {1cm, 0cm, 1cm, 1cm}, clip, width=0.8\textwidth]{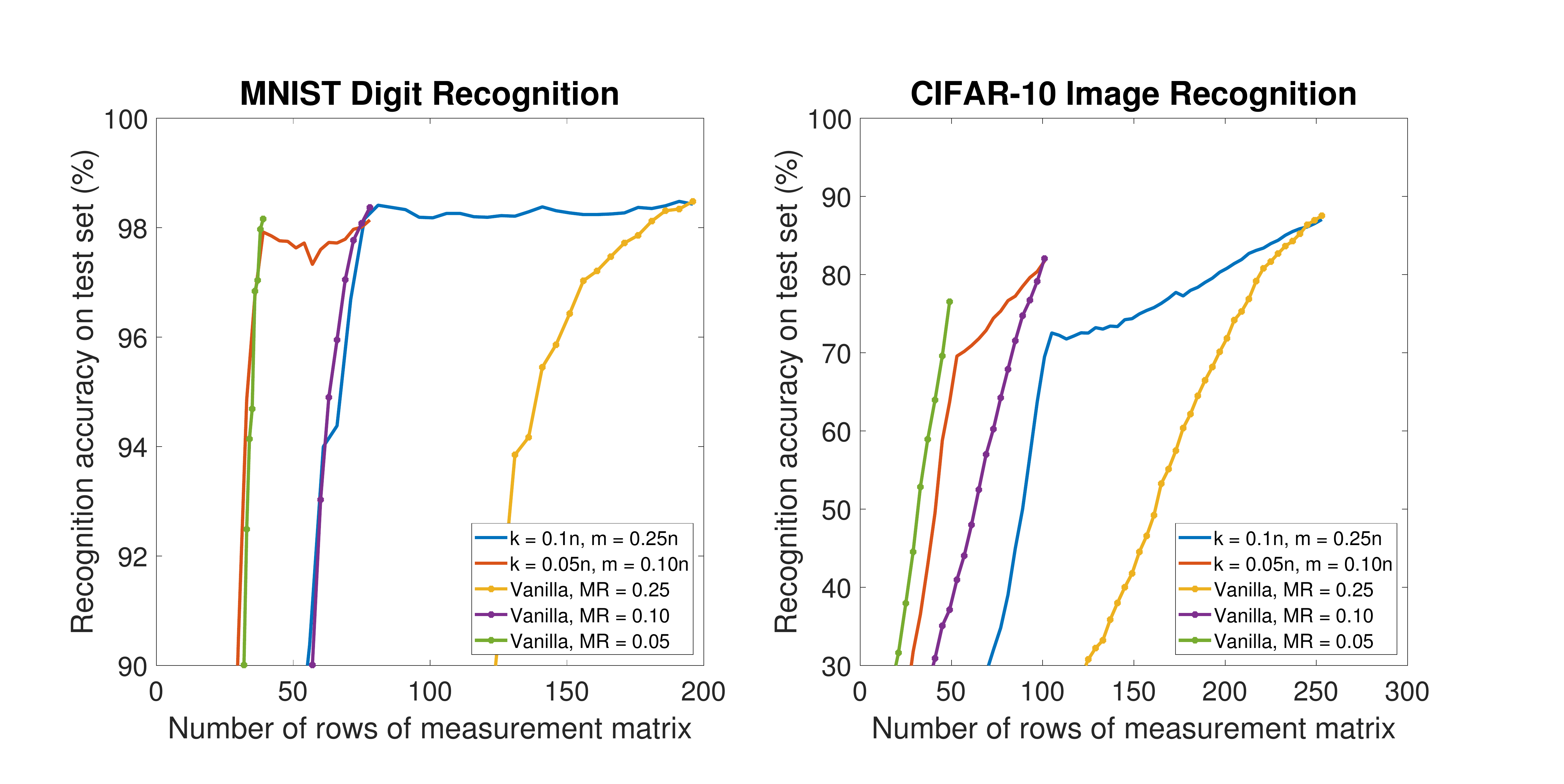}
\caption{Rate-Adaptive CS results for image recognition for MNIST hand-written digit recognition ($n=784$) and CIFAR-10 object recognition ($n=1024$) -- compared to the vanilla training algorithm trained for a single MR. Clearly, Rate-Adaptive CS is stable over the chosen operating range compared to the vanilla framework. Best viewed in color.}
\label{fig:stable}
\end{figure*}


\subsubsection{Results:} Table \ref{table:stable_mnist} Figure \ref{fig:stable} shows comparison between Rate-Adaptive CS and vanilla training algorithm for different variations of $[k,m]$. We observe that the performance of Rate-Adaptive CS decreases slowly as the number of measurements decreases within the operating range for both MNIST and CIFAR-10. This is contrary to the performance of vanilla training algorithm where the performance falls more steeply as we move away from training MR. For the case of CIFAR-10, we note that a rate-adaptive network for $MR = [0.10,0.25]$ performs on average 8.49\% points and up to 29.12 \% points better than a vanilla network trained for $MR = 0.25$, when tested over all MRs.

\begin{table*}
\normalsize
\centering
\caption{Comparison of Rate-Adaptive CS algorithm with Vanilla training algorithm on MNIST ($n=784$) and CIFAR-10 ($n=1024$) datasets in terms of recognition accuracy (\%) on the test set. The vanilla algorithm trains the network for a single MR, while Rate-Adaptive CS is trained so that, at test time, the system is stable over the entire chosen MR range. This can be easily observed from the results shown}
\begin{tabular}{|c|c|p{0.8cm}|p{0.8cm}|p{0.8cm}|p{0.8cm}|p{0.8cm}|p{0.8cm}|}
\hline
\multirow{2}{*}{Method} & \multirow{2}{*}{\makecell{Training\\MR}} & \multicolumn{3}{c|}{\makecell{MNIST LeNet\\Test MR}} & \multicolumn{3}{c|}{\makecell{CIFAR-10 ResNet\\Test MR}}\\
\cline{3-8}
& & 0.25 & 0.10 & 0.05 & 0.25 & 0.10 & 0.05 \\
\hline
Vanilla & 0.25 & 98.48 & 66.13 & 28.86 & 87.53 & 21.1 & 13.9\\
\hline
Vanilla & 0.10 & N/A & 98.37 & 69.57 & N/A & 82.1 & 40.0\\
\hline
Vanilla & 0.05 & N/A & N/A & 98.16 & N/A & N/A & 76.53\\
\hline
Rate-Adaptive CS & $0.10 - 0.25$ & 98.43 & 98.34 & 77.41 & 87.05 & 72.7 & 21.0\\
\hline
Rate-Adaptive CS & $0.05 - 0.10$ & N/A & 98.14 & 97.92 & N/A & 81.72& 69.1\\
\hline
\end{tabular}
\label{table:stable_mnist}
\end{table*}

\section{Conclusion}
In this paper, we design a novel training algorithm that enables training a single network that can be operated over a range of measurement rates, thus overcoming a major drawback of previous related algorithms. Our rate-adaptive framework performs significantly better than previous algorithms that work only for a single measurement rate. We demonstrate this on two important problems -- image reconstruction and image recognition. Furthermore, through object tracking, we have shown how the rate-adaptive framework can be utilized in systems with time-varying constraints on the measurement rate. Future work includes extending these ideas to other network architectures and application domains such as multi-task learning and metric learning. We hope that the algorithm and results presented in this paper will enable researchers to adopt spatial multiplexing/  compressive imaging more easily in real-world scenarios.


%



%

\bibliographystyle{IEEEtran}

\bibliography{IEEEabrv,egpaper_final}

%

\begin{IEEEbiography}{Suhas Lohit}
(S’15) is a PhD student in Electrical Engineering at Arizona State University, Tempe, USA. He received his MS in Computer Engineering also at Arizona State University in 2015. His research interests are in computational imaging, computer vision and deep learning. He was awarded the University Graduate Fellowship in 2015 and the Best Paper Award at the Computational Cameras and Devices Workshop held in conjunction with CVPR 2015.
\end{IEEEbiography}

\begin{IEEEbiography}{Rajhans Singh}
is a PhD student in Electrical Engineering at Arizona State University, Tempe, USA. He received his B. Tech degree in Electrical Engineering from the Indian Institute of Technology Rookie, India, in 2015. His research interests are in computer vision, deep learning and generative models.
\end{IEEEbiography}

\begin{IEEEbiography}{Kuldeep Kulkarni}
is a post-doctoral research associate in the department of electrical and computer engineering at Carnegie Mellon University. He received the B.Tech degree in electrical and electronics engineering from the National Institute of Technology Karnataka, Surathkal, India, in 2009. Later, he received the M.S degree in electrical engineering and the PhD degree in electrical engineering with arts, media and engineering (AME) concentration from Arizona State University in 2012 and 2017 respectively.  His research interests lie at the intersection of the areas of computer vision, machine learning and compressive sensing. He received the best paper award in the Computational Cameras and Displays (CCD) Workshop 2015, held in conjunction with the IEEE CVPR conference.
\end{IEEEbiography}

\begin{IEEEbiography}{Pavan Turaga}
(S’05, M’09, SM’14) is Associate Professor in the School of Arts, Media, Engineering, and Electrical Engineering at Arizona State University. He received the B.Tech. degree in electronics and communication engineering from the Indian Institute of Technology Guwahati, India, in 2004, and the M.S. and Ph.D. degrees in electrical engineering from the University of Maryland, College Park in 2008 and 2009 respectively. He then spent two years as a research associate at the Center for Automation Research, University of Maryland, College Park. His research interests are in computer vision and computational imaging with a focus on non-Euclidean and high-dimensional statistical techniques for these applications. He was awarded the Distinguished Dissertation Fellowship in 2009. He was selected to participate in the Emerging Leaders in Multimedia Workshop by IBM, New York, in 2008. He received the National Science Foundation CAREER award in 2015.
\end{IEEEbiography}




\end{document}